\documentclass[journal]{IEEEtai}

\usepackage[colorlinks,urlcolor=blue,linkcolor=blue,citecolor=blue]{hyperref}

\usepackage{color,array}

\usepackage{graphicx}
\usepackage{wrapfig}
\usepackage{graphicx}
\usepackage[space]{grffile}
\usepackage{latexsym}
\usepackage{textcomp}
\usepackage{longtable}
\usepackage{tabulary}
\usepackage{booktabs,array,multirow}
\usepackage{amsfonts,amsmath,amssymb}
\usepackage{tabularray}

\begin{document}

\title{Fair Machine Learning in Healthcare: A Survey}

\author{Qizhang Feng, Mengnan Du, Na Zou, and Xia Hu
\thanks{Manuscript submitted June 17, 2022; date of current version Nov 7, 2023. This work is in part supported by NSF grants IIS-1939716 and IIS-1900990.}
\thanks{Qizhang Feng is with the Department of Computer Science \& Engineering, Texas A\&M University, TX 77843, US (e-mail: qf31@tamu.edu).}
\thanks{Mengnan Du is with the Department of Data Science, New Jersey Institute of Technology, NJ 07102, US (e-mail: mengnan.du@njit.edu).}
\thanks{Na Zou is with the Department of Engineering Technology \& Industrial Distribution, Texas A\&M University, TX 77843, US (e-mail: nzou1@tamu.edu).}
\thanks{Xia Hu is with the Department of Computer Science, Rice University, TX 77251, US (e-mail: xia.hu@rice.edu).}
\thanks{This paragraph will include the Associate Editor who handled your paper.}}

\markboth{Journal of IEEE Transactions on Artificial Intelligence, Vol. 00, No. 0, Month 2020}
{First A. Author \MakeLowercase{\textit{et al.}}: Bare Demo of IEEEtai.cls for IEEE Journals of IEEE Transactions on Artificial Intelligence}

\maketitle

\begin{abstract}
The digitization of healthcare data coupled with advances in computational capabilities has propelled the adoption of machine learning (ML) in healthcare. However, these methods can perpetuate or even exacerbate existing disparities, leading to fairness concerns such as the unequal distribution of resources and diagnostic inaccuracies among different demographic groups. Addressing these fairness problem is paramount to prevent further entrenchment of social injustices.
In this survey, we analyze the intersection of fairness in machine learning and healthcare disparities. We adopt a framework based on the principles of distributive justice to categorize fairness concerns into two distinct classes: equal allocation and equal performance. We provide a critical review of the associated fairness metrics from a machine learning standpoint and examine biases and mitigation strategies across the stages of the ML lifecycle, discussing the relationship between biases and their countermeasures.
The paper concludes with a discussion on the pressing challenges that remain unaddressed in ensuring fairness in healthcare ML, and proposes several new research directions that hold promise for developing ethical and equitable ML applications in healthcare.
\end{abstract}

\begin{IEEEImpStatement}

Along with the rapid growth in the use of machine learning in healthcare in recent years, there has been a growing concern about the fairness problems that come along with it.
This survey article helps break down the barriers between fair machine learning and healthcare, and aims to: 1) improve healthcare practitioners' understanding of the bias of machine learning in healthcare from a computational perspective; 2) assist machine learning researchers in establishing a clear picture on how to develop fair algorithms in various healthcare scenarios from a healthcare perspective; and 3) increase public trust in machine learning algorithms and promote the use of machine learning methods in real-world healthcare settings.
\end{IEEEImpStatement}


\section{Introduction}

\IEEEPARstart{W}{ith} the advent of sophisticated machine learning (ML) applications in healthcare, from medical image analysis to electronic health records processing, we stand on the cusp of a transformative era in medicine~\cite{Kwak2019DeepHealthRA,ronneberger2015u,Scherrer2012SuperresolutionRT,Xu2014DeepLO,Liu2020SemiSupervisedMI}. Despite these advancements, there remains a significant yet understudied challenge: ensuring fairness in algorithmic decisions, particularly as they relate to the equitable treatment of diverse patient populations~\cite{xu2022algorithmic}.

Fairness in healthcare ML refers to the equitable distribution of benefits and burdens across all demographic groups, with particular attention to historically marginalized communities. It encompasses a range of issues, from the allocation of healthcare resources to diagnostic accuracy across different patient demographics. Notable instances include genetic tests where AI models disproportionately misrepresent risks for minority groups~\cite{Parikh2019AddressingBI}, and diagnostic discrepancies exacerbated by incomplete medical records among Black and Hispanic patients~\cite{SeyyedKalantari2021CheXclusionFG}. The Covid-19 pandemic has further highlighted these disparities, intensifying the urgency to address them~\cite{giovanola2022beyond}.

{\color{blue}
Recognizing the potential of ML to either perpetuate or mitigate existing disparities, this survey seeks to fill the critical gap in the literature by providing a comprehensive analysis of fairness-oriented ML strategies in healthcare. We acknowledge the socio-technical nature of fairness challenges in healthcare ML, which encompasses algorithmic aspects and extends to societal, ethical, and regulatory dimensions. This survey synthesizes insights from previous works, including the categorization of fairness problems~\cite{Rajkomar2018EnsuringFI} and solutions, and charts a path forward for equitable ML applications in healthcare. The commitment to fair and inclusive AI development is echoed by governmental bodies, such as the National Institutes of Health, through initiatives like AIM-AHEAD and Bridge2AI~\cite{brogan2021next}.
}


\noindent\textbf{Distinguishing from related reviews.}
While the domains of fairness in machine learning and health disparities are well-researched, their intersection remains nascent. Several surveys~\cite{Rajkomar2018EnsuringFI, Chen2021AlgorithmFI, fletcher2021addressing, Xu2022AlgorithmicFI,chen2021ethical,ricci2022addressing} have attempted to address fairness problems in machine learning methods for healthcare. Yet, a holistic approach that captures both the ethical and technical detail is missing in the literature.
The key point of fair machine learning in healthcare contain both ethical consideration and also technical details.
However, We have observed that existing works tend to focus on one aspect while neglecting the other. For instance, some works~\cite{Rajkomar2018EnsuringFI, chen2021ethical} discuss fairness from an ethical standpoint but lack a detailed connection with technical mitigation methods and metrics for fairness in machine learning. 
Conversely, another line of work~\cite{Xu2022AlgorithmicFI} provides a technical perspective by categorizing fairness metrics and mitigation methods but does not establish a strong link with the ethical aspects of healthcare fairness. 
Additionally, some studies~\cite{Chen2021AlgorithmFI} focus narrowly on data shifts and federated learning, while work~\cite{ricci2022addressing} limits its scope to fairness in artificial intelligence for medical imaging. Our survey seeks to establish a comprehensive link between the ethical and technical dimensions of fair machine learning in healthcare.
Our survey is motivated by the need to bridge this evident gap, providing a comprehensive perspective that ties together the ethical considerations and technical details of fairness in healthcare machine learning.
Specifically, our contributions are summarized as follows:
\begin{enumerate}
    \item Connect the ethical and technical aspects of fairness in healthcare by adopting the concept of distributive justice from works~\cite{Rajkomar2018EnsuringFI,kuppler2021distributive}. Classify healthcare fairness problems into equal allocation and equal performance, and provide a comprehensive summary of fairness measurement methods within the fair machine learning domain, categorizing them accordingly.
    \item Provide a comprehensive overview of biases at various stages of the machine learning model development. Conduct a structured analysis of fairness mitigation methods, surpassing previous surveys in exhaustiveness. Highlight the critical gap in current mitigation methods, focusing on the necessity to discuss and analyze their applicability to scenarios of equal allocation and equal performance.
    \item Discuss challenges and opportunities in creating a fair and reliable machine learning ecosystem for healthcare, with an emphasis on the unique aspects of healthcare applications.
\end{enumerate}

The structure of this article is listed as follows. The definition of fairness problems in healthcare is given in Section~\ref{sec:ml-health}. On the basis of this definition, measurements of fairness are given in Section~\ref{sec:measure}.
Then the biases at various stages of model development are introduced in Section~\ref{sec:source}. Similarly, according to the same categorization, methods for mitigating fairness problems are discussed in Section~\ref{sec:mitigate}.
Finally, we highlight the challenges and opportunities for a fair and trustworthy machine learning healthcare ecosystem based on uniqueness in healthcare applications in Section~\ref{sec:challenge}.

\section{Fairness Problems in ML for Healthcare }\label{sec:ml-health}
Because of the digitization of medical data collection, we can now collect large amounts of medical data and develop machine learning algorithms for a variety of medical tasks.
First, machine learning models have been used in pioneering applications on medical images (e.g., NIH Chest-Xray14, CheXpert, MIMIC-CXR and Chest-Xray8~\cite{Wang2017ChestXRay8HC, Irvin2019CheXpertAL, Johnson2019MIMICCXRAL, SeyyedKalantari2021CheXclusionFG}). For example, a large-scale study built a deep neural network on the NIH Chest-XRay14 dataset and the CheXpert dataset to diagnose various chest diseases~\cite{Larrazabal2020GenderII}.
Second, machine learning models have also been applied to the structured electronic health record (EHR), which contains information on demographics, diagnoses, laboratory tests, medications, etc. For instance, a gradient boosting model was used to predict cardiovascular disease risk based on the Stanford Translational Research Integrated Database Environment (STRIDE 8) dataset~\cite{Nguyen2019PredictingCR}.
Third, advancements in natural language processing (NLP) have greatly enhanced our ability to process unstructured electronic health record (EHR) data, such as clinical narratives, medical examinations, clinical laboratory reports, surgical notes, and discharge summaries. These NLP methods facilitate a range of critical tasks including medical concept extraction, disease inference, and clinical decision support~\cite{choi2016learning, Zhang2020HurtfulWQ, Minot2021InterpretableBM}. Particularly, large language models (LLMs) such as GPT-2 and GPT-3 have demonstrated their utility in medical question-answering tasks, including those in pain management domains~\cite{Radford2019LanguageMA,Brown2020LanguageMA}.
{\color{blue}
The recent surge in conversational language models, exemplified by ChatGPT, underscores their transformative potential in healthcare. Recent study evaluates the feasibility of ChatGPT across multiple clinical and research scenarios, showcasing the model's substantial impact and the breadth of its applications in the healthcare field~\cite{cascella2023evaluating}.
The recent surge in conversational language models, such as ChatGPT, marks a significant shift in healthcare technology. Notably, Microsoft's Azure Health Bot~\cite{AzureHea70:online} and NHS-LLM~\cite{ALargeLa55:online} represent pioneering applications of LLMs in healthcare. These models are designed to assist users in assessing healthcare needs, particularly when they are uncertain about the severity of their condition. However, this raises critical fairness considerations, as the algorithms' decision-making processes must account for diverse patient populations and their unique healthcare requirements. This is crucial to ensure equitable access and outcomes across different demographics, thus avoiding exacerbation of existing disparities in healthcare~\cite{cascella2023evaluating}.
}

\subsection{Distributive Justice in Machine learning for Healthcare}\label{justice}
Although the use of machine learning techniques in healthcare has been shown to correct clinical inadequacies and increase operational efficiency by reducing resource waste~\cite{topol2019high}, various fairness problems have also been raised.
The problem of fairness in machine learning methods is reflected in the discrimination of different groups~\cite{giovanola2022beyond}.
For example, state-of-the-art convolutional neural network (CNN) classifiers were found to differ in the true positive rate across protected attributes (e.g., patient gender, age, race, and insurance type) on 14 diagnostic tasks in 3 well-known public chest x-ray datasets~\cite{SeyyedKalantari2021CheXclusionFG}.

Discrimination can be understood as a distributive problem~\cite{lamont2017distributive}.
In studies developed related to machine learning in healthcare, fairness problems often refer to the unequal distribution of resources such as medical care, clinical services and health facilities~\cite{friedler2016possibility}.
Distributive justice is concerned with the distribution of resources among members of a society, and the underlying idea of distributive justice theories are distribution principles and metrics of justice~\cite{kuppler2021distributive}.
The distribution principles specify how resources should be distributed~\cite{lamont2017distributive}.
The justice metric specifies the type of resources to be allocated~\cite{kuppler2021distributive}.
In the context of the fairness problem of machine learning methods in healthcare, resources often refer to the medical services allocated by the system, or the error rate of the predictions it gives.
Fairness problems can be grouped into two categories based on differences in the resources allocated: equal allocation and equal performance~\cite{Rajkomar2018EnsuringFI}.

\subsubsection{Equal Allocation}
Machine learning models are often used to allocate medical supplies such as vaccines, medicines and organ transplants. 
Accordingly, fairness problems of machine learning methods in healthcare occur if the model determines that the allocation of resources is not equal between groups.
For example, a recently published work focused on building models to help determine which patients with chronic kidney disease should undergo kidney transplantation.
The study found that the model was biased towards black patients and tended to classify black patients as having more severe kidney disease~\cite{ahmed2021examining}.
Another study examined how to effectively combat influenza in the early stages of an influenza outbreak with minimal vaccine dosing in the early stages of vaccine production when production is limited~\cite{enayati2020optimal}. They found that the optimal solution to the model was likely to produce a controversial distribution strategy of not distributing any vaccine to certain subgroups.

\subsubsection{Equal Performance}
In some medical situations, machine learning models are used for medical tasks such as disease diagnosis, mortality prediction and multi-organ segmentation, etc.
Consequently, it would be unfair if the performance and results of machine learning models are not equally accurate in terms of metrics such as accuracy for patients in different demographic groups.
A recent study discovered that, despite having similar accuracy to board-certified dermatologists, machine learning algorithms used to classify images of benign and malignant moles are less accurate in the diagnostic task of melanoma on dark skin~\cite{adamson2018machine}.
Another study analyzes the sex/racial bias in AI-based cine CMR segmentation using a large-scale database~\cite{PuyolAntn2021FairnessIC}.
It is shown that state-of-the-art deep learning models for automatic segmentation of the ventricle and myocardium based on cine short-axis CMR had statistically significant differences in 
errors between races.

Different healthcare settings require different kinds of distributive justice. 
The various distributive justice options make it extremely difficult for ML models to satisfy all conditions~\cite{dieterich2016compas, chouldechova2017fair}. Thus, suitable metrics are crucial for evaluating the fairness of a machine learning model. 

\begin{figure*}[h]
    \centering
    \includegraphics[width=1.0\linewidth]{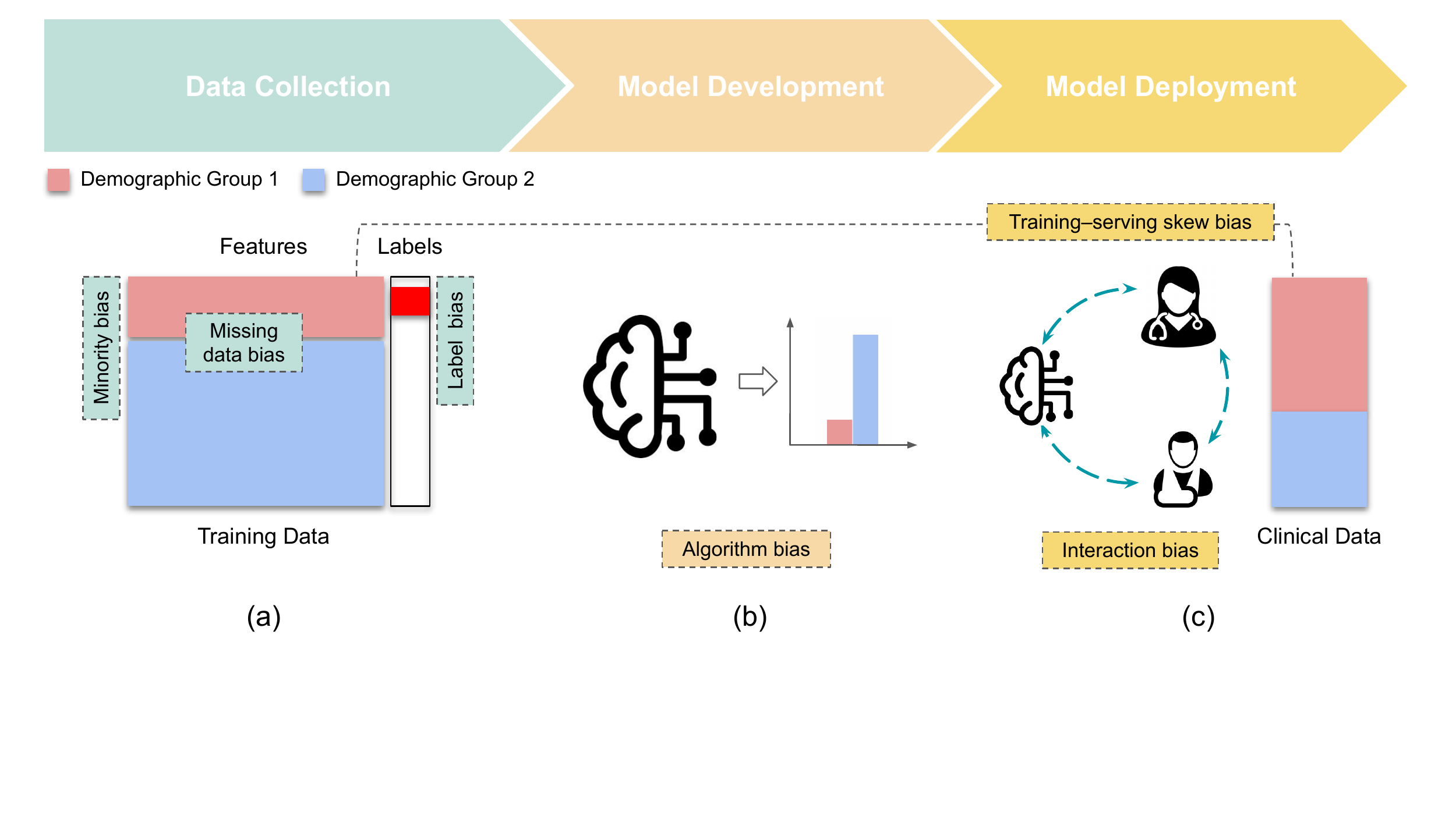}
    
    \vspace{-70pt}
    \caption{\textbf{Bias at the different stages in machine learning systems}: Red and blue represent two demographic groups. (a) The biases that exist at the data collection stage include minority bias, missing-data bias and label bias. Minority bias occurs when the sample size of the demographic groups are unbalanced.  Missing data bias occurs when data may be missing in a non-random way. Label bias occurs when the quality of labels varies between different demographic groups. (b) Algorithm bias exists in model development stage, leads to systematical unfair results for certain demographic group. (c) The biases that exist at the data collection stage include interaction bias and training-serving skew bias. Training-serving skew bias occurs when the distribution of data in the deployment stage differs from the distribution of data in the training phase. Interaction bias occurs patients and healthcare professionals interact with machine learning models. Please refer to section~\ref{sec:source} for further details. }
    \label{fig:pipe}
\end{figure*} 

\section{Measurement of Fairness}\label{sec:measure}
In the previous section, we introduce the fairness problems in machine learning for healthcare and categorize them into equal allocation problems and equal performance problems.
Selecting an appropriate fairness metric is critical to measuring the fairness problem in various healthcare scenarios.
In this section, we first introduce two principles of fairness following distributive justice and then summarize the common metrics of fairness that apply to them.

To measure the fairness of a given decision algorithm $f(\cdot)$, we define $\mathbf{x} \in \mathbb{R}^{d_x}$ as the nonsensitive features vector and $\mathbf{z} \in \mathbb{R}^{d_z}$ as the sensitive features vector. In most cases, only one sensitive feature is considered, so we use $z$ when $d_z = 1$. The prediction of the model $f(\cdot)$ with input $\mathbf{x}$ as $\hat{y} = f(\mathbf{x})$, and $y$ is the corresponding ground truth label. 
In this survey, we mainly focus on the binary classification problem, while many works go beyond it into multi-class classification task, regression task, segmentation task with their own unique metric.
Other symbols and definitions can be found in Table~\ref{tab:symbols}.

\begin{table*}[]
    \centering
    \caption{Main symbols and definitions.}
    \begin{tabular}{l|l} \toprule
    \textbf{Symbol} & \textbf{Definition} \\
    \midrule
    $f(\cdot)$ & A machine learning model that maps attributes to predictions. \\
    $\textbf{x} \in \mathbb{R}^{d_x}$ & The non-sensitive attributes with a dimension of $d_x$.  \\
    $\textbf{z} \in \mathbb{R}^{d_z}$ & The sensitive attribute.  \\
    $z$ & The sensitive attribute when ${d_z}=1$.  \\
    $\hat{y} \in \{0, 1\}$ & A binary prediction that indicates negative and positive outcomes for $0$ and $1$, respectively.  \\
    $y \in \{0, 1\}$ & A binary ground truth that indicates negative and positive outcomes for $0$ and $1$, respectively. \\
    $\hat{y}_{\{\mathrm{z} \leftarrow \mathrm{a}\}}$ & A prediction in the counterfactual world if $z = a$.  \\
    $\mathcal{D}$ & The training dataset.  \\
    $d(\cdot, \cdot)$ & Distance of two individuals in the attribute space.  \\
    $D(\cdot, \cdot)$ & Distance of two individuals in the prediction space.  \\
    $\theta$ & Parameters $\theta$ of the backbone network. \\
    $\phi$ & Parameters $\phi$ of the adversarial network. \\
    $A_{\phi}(\cdot)$ & Adversarial network with  parameters $\phi$ \\
    $L(\mathcal{D};\theta)$ & The downstream task loss. \\
    $L_{adv}(\mathcal{D};\phi)$ & The adversarial loss. \\

    \bottomrule
    \end{tabular}
    \label{tab:symbols}
\end{table*}

\subsection{Equal Allocation}
Equal allocation is suitable in the healthcare setting when resources should be distributed proportionally to patients in protected groups.
Equal allocation is also applicable when the label is historically biased~\cite{rajkomar2018ensuring}. For example, if historically African American women have been sent for such procedures at unduly low rates, then a `correct' prediction based on historical data would underestimate the status of these women.
From a computational point of view, it is desirable that the decisions made by the model differ as little as possible between the different demographic groups.
In the following, we introduce some fairness metrics that follow the principle of equal allocation.
\begin{itemize}
    \item \textbf{Demographic Parity (DP)} is satisfied if a machine learning algorithm gives equal decision rates 
    for different demographic subgroups a and b:
    \begin{equation}
        \mathbb{P}(\hat{y}=1 \mid z=a)=\mathbb{P}(\hat{y}=1 \mid z=b),
    \end{equation}
    DP can be extended for multi-class classification application such as image recognition, text categorization, etc~\cite{Denis2021FairnessGI}:
    \begin{equation}
        \sum_{k=1}^{K}\left| \mathbb{P}(\hat{y}=k|z=a) -  \mathbb{P}(\hat{y}=k|z=b) \right|=0, \;\; \forall k \in [K],
    \end{equation}
    where $[K] = \{1,\dots, K\}$ indicates $K$ number of classes. An alternative definition can constitute the summation with a maximum. DP can also be applied to the regression model rather than to the classification model~\cite{Chzhen2020FairRW}:
    \begin{equation}
        \sup_{t\in \mathbb{R}}\left| \mathbb{P}(\hat{y}\leq t|z=a) - \mathbb{P}(\hat{y}\leq t|z=b) \right|=0.
    \end{equation}
    
    \item \textbf{General Demographic Parity (GDP)}~\cite{jiang2021generalized} extends the demographic parity on the continuous sensitive attribute:
    \begin{equation}
    \Delta G D P=\mathbb{E}_{z}\left[\left|\mathbb{E}[\hat{y} \mid z]-\mathbb{E}[\hat{y}]\right|\right],
    \end{equation}
    where $\mathbb{E}[\hat{y} \mid s]$ is the local average prediction of the model conditioned on the sensitive attribute, and $\mathbb{E}[\hat{y}]$ is the global prediction average.
    GDP degenerates into weighted demographic parity for the categorical sensitive attribute.
    
    \item \textbf{Fairness through Unawareness (FTU)}~\cite{Kusner2017CounterfactualF} defines an algorithm as FTU fair as long as sensitive attributes are not used by the decision-making algorithm $f(\cdot)$:
    \begin{equation}
        \mathbb{P}(\hat{y} \mid \mathbf{x}, z)=\mathbb{P}(\hat{y} \mid \mathbf{x})
    \end{equation}
    FTU will fail even if no sensitive attributes are present in the data, if a combination of non-sensitive features can act as a proxy for them. For example, an individual's postal code might be used as a proxy for their income, race, or ethnicity ~\cite{fabris2023measuring}.
    
    \item \textbf{Fairness through Awareness}~\cite{Dwork2012FairnessTA} emphasizes that a fair algorithm should make similar decisions for two individuals $x$ and $x'$ with similar non-sensitive attributes:
    \begin{equation}
        D\left(f(x), f(x')\right) \leq d\left(x, x'\right)
    \end{equation}
    Note that the algorithm should satisfy the $(D, d) \text {-Lipschitz }$ property.
    
    \item \textbf{Counterfactual Fairness}~\cite{Kusner2017CounterfactualF} is derived from causal theory. The intuition of counterfactual fairness is that a fair algorithm should provide the same decision for a real-world individual and its corresponding one in the counterfactual world:
    \begin{equation}
        \mathbb{P}\left[\hat{y}_{\{\mathrm{z} \leftarrow \mathrm{a}\}}=\mathrm{c} \mid \mathrm{x}, \mathrm{z}=\mathrm{a}\right]=\mathbb{P}\left[\hat{y}_{\{\mathrm{z} \leftarrow b\}}=\mathrm{c} \mid \mathrm{x}, \mathrm{z}=\mathrm{a}\right]
    \end{equation}
    Achieving consensus on causal graphs is challenging due to the complexity of causal structure discovery, particularly without existing knowledge of causality. This complexity can lead to the incorrect assumption of causal structures from statistical model outputs, resulting in varying interpretations and difficulty in standardizing causal graphs~\cite{shen2020challenges}.

\end{itemize}

\subsection{Equal Performance}
Equal performance means that a model is guaranteed to be equally accurate for patients in protected and non-protected groups. The concept of accuracy can include equal sensitivity (also called equal opportunity~\cite{Zafar2017FairnessBD}), equalized odds, and equal positive predictive value~\cite{chouldechova2017fair}, or broader metrics such as AUC, etc.
Equal performance metric is appropriate in the context where the accuracy of the machine learning model is crucial.
For instance, the machine learning system can be introduced to build a monitoring system that is used to alert rapid response teams when hospitalized patients are at high risk of deterioration~\cite{Escobar2016PilotingEM}.
If the predictive model imposes a high false positive rate on the protected group, patients in the protected group may lose the opportunity to be identified, which can have serious consequences.
However, forcing a model's predictions to have one of the performance characteristics of equality~\cite{hardt2016equality} may have unintended consequences. For example, the model may achieve equal odds by sacrificing the accuracy of the unprotected group, which undermines the benefit principle~\cite{wang2022brief}.
In the following, we introduce some fairness measurements that follow the principle of equal performance.
\begin{itemize}
    \item \textbf{Equal Opportunity} is preferred when people care more about true positive rates. We say that a classifier satisfies equal opportunity if the true positive rate is the same across the groups~\cite{Zafar2017FairnessBD}:
    \begin{equation}
        \mathbb{P}\{\hat{y}=1 \mid y=1, z=a\}=\mathbb{P}\{\hat{y}=1 \mid y=1, z=b\}.
    \end{equation}
    It can also be referred to as positive predictive value parity. Similarly, there is negative predictive value parity:
    \begin{equation}
        \mathbb{P}\{\hat{y}=1 \mid y=0, z=a\}=\mathbb{P}\{\hat{y}=1 \mid y=0, z=b\}
    \end{equation}
    Predictive value parity is also called sufficiency.

    \item \textbf{Equalized Odds} requires that the decision rates across demographic subgroups be the same when their outcome is the same~\cite{Caton2020FairnessIM}:
\begin{multline}
    \mathbb{P}\{\hat{y}=1 \mid z=a, y=0\} = \mathbb{P}\{\hat{y}=1 \mid z=b, y=0\}, \\
    \mathbb{P}\{\hat{y}=1 \mid z=a, y=1\} = \mathbb{P}\{\hat{y}=1 \mid z=b, y=1\}
\end{multline}
Equalized Odds requires the algorithm to have equal true positive rates and equal false positive rates at the same time.
    \item \textbf{Treatment Equality} requires that the ratio of false negatives and false positives be the same for subgroups~\cite{Berk2018FairnessIC}:
    \begin{equation}
        \frac{\mathbb{P}(\hat{y}=1|y=0, z=a))}{\mathbb{P}(\hat{y}=0|y=1, z=a))} = \frac{\mathbb{P}(\hat{y}=1|y=0, z=b))}{\mathbb{P}(\hat{y}=0|y=1, z=b))}.
    \end{equation}
\end{itemize}

\section{Sources of Fairness Problems}\label{sec:source}

In this section, we summarize the causes of fairness problems in healthcare machine learning and use the term `bias' to denote them~\cite{Mehrabi2021ASO}. 
The process of building a machine learning-based healthcare system can be divided into three stages. First, the agency collects relevant clinical data for model development. Then, developers select and train a suitable model for the intended task, based on the data and the type of task. Finally, the institution involved can license the model for implementation in real clinical practice.
We present the various complex biases that exist in healthcare based on the three different stages (see the overview in Figure~\ref{fig:pipe}).

\subsection{Bias in Data Collection}\label{sec_bias_in_dc}
Data collection is the first stage at which bias may be introduced.
A machine learning model is trained to fit the distribution of the training data.
{\color{blue}
When there is bias in the data, the model may perpetuates the bias}
(as shown in Figure~\ref{fig:pipe}.(a)).
In the following paragraphs, we review several common types of data bias in clinical practice.

\subsubsection{Minority Bias}
Minority bias occurs when the sample size of a demographic group is smaller than that of other groups.
The development of machine learning algorithms in healthcare is currently highly dependent on public biobank databases~\cite{EhteshamiBejnordi2017DiagnosticAO, Campanella2019ClinicalgradeCP}. However, due to the uneven development of medical standards, most of the data collection is done in Europe. This has led to the study of human knowledge of the disease using biobank repositories that mainly represent individuals of European ancestry.
For example, the vast majority of cases in the Cancer Genome Atlas (TCGA) are made up of whites, representing approximately 82.0\% of the cases. In contrast, a very small proportion of the cases are from black, Asian, and other ethnic minorities~\cite{Gao2013IntegrativeAO}.
In fact, demographic data such as ethnicity are crucial to determining the mutational profile and mechanisms of cancer. As a result, genetic risk models perform worse in ethnic minority populations.

\subsubsection{Missing-data Bias}
Missing data bias occurs when data may be missing in a non-random way. Machine learning algorithms may cause harm to people with missing data in the dataset.
For example, research has found that vulnerable people of low socioeconomic status are likely to be seen in a piecemeal fashion or cannot be seen. If patients are identified based on a certain number of ICD codes, records of the same number of visits to several different healthcare systems for these patients may be missing.
Another example is that, despite numerous initiatives, sexual orientation and gender identity have been largely absent from electronic health records to date.
Machine learning-based clinical decision support systems can misinterpret the lack of access to care as a lower burden of disease and therefore produce inaccurate predictions for these groups~\cite{Chen2021AlgorithmFI}.


\subsubsection{Label Bias}
Label bias may also be present in data labels, and the quality of the labels can contribute to bias~\cite{Rajkomar2018EnsuringFI}. For example, people with low socioeconomic status may be more likely to be seen in teaching clinics, where documentation or clinical reasoning may be less accurate or systematically different from the care provided to patients with high socioeconomic status. Algorithms based on these data may reflect practitioner bias and misclassify patients based on these factors.
The choice of inappropriate labels can also introduce bias. For example, some models use specific phrases that appear in clinical records as proxy labels that indicate the presence of cardiovascular disease. However, because women have different symptoms of acute coronary syndromes, proxy phrases have different meanings for men and women. As a result, women can receive delayed care, which causes discrimination against women.

\subsection{Bias in Model Development}\label{sec:bias_in_mdev}
Bias in the model development phase can lead to machine learning models perpetuating or even amplifying existing biases in the data. This can stem from various sources, including inappropriate intrinsic hypotheses, the structure of the model, and biased loss estimators, all of which can potentially contribute to fairness problems~\cite{carbonell1983overview, Williams2019TowardsQO, chen2018my}. A predominant concern in this context is \textit{algorithmic bias}, where the source of bias is traceable back to the model itself, systematically leading to unfair results for certain groups as depicted in Figure~\ref{fig:pipe}.(b).

\begin{table*}[t]
\fontsize{4.6}{6.8}\selectfont
\centering
\caption{Mitigation Methods Categorization.}
\label{fig:miti_cat}
\begin{tblr}{
  cell{2}{1} = {r=9}{},
  cell{2}{2} = {r=7}{},
  cell{2}{3} = {r=2}{},
  cell{4}{3} = {r=2}{},
  cell{7}{3} = {r=2}{},
  cell{9}{2} = {r=2}{},
  cell{11}{1} = {r=7}{},
  cell{11}{2} = {r=6}{},
  cell{11}{3} = {r=4}{},
  cell{18}{1} = {r=3}{},
  hline{1,21} = {-}{0.08em},
  hline{2,11,18} = {-}{0.05em},
  hline{4,6-7,15-16} = {3-7}{0.03em},
  hline{9,17,19-20} = {2-7}{0.03em},
}
                  &                       &                        & Reference                                  & Task                                                                    & Dataset                                                                                                                                       & Data Type          \\
Data Collection   & Data Redistribution   & Diversified Collection & \cite{li2023targeting}           & Type II Diabetes Risk Prediction                                        & eMERGE\cite{mccarty2011emerge}                                                                              & EHR,Genomic        \\
                  &                       &                        & \cite{Fan2021OnTF}               & Skin Lesion Classification                                              & Skin ISIC 2018\cite{codella2018skin}                                                                        & Medical Image      \\
                  &                       & Data Reweighting       & \cite{Xue2019RobustLA}          & Skin Lesion Classification                                              & Skin ISIC 2017\cite{Gutman2018SkinLA}                                                                     & Medical Image      \\
                  &                       &                        & \cite{Wachinger2016DomainAF}     & AD Classification                                                       & ANDI\cite{petersen2010alzheimer}                                                                            & Medical Image      \\
                  &                       & Data Resampling        & \cite{Chawla2002SMOTESM}         & Diabetes Classification                                                 & The Pima Indian Diabetes Dataset\cite{smith1988using}                                                       & EHR                \\
                  &                       & Synthetic Data         & \cite{rajotte2021reducing}       & Skin Lesion Classification                                              & HAM10000\cite{Tschandl2018TheHD}                                                                            & Medical Image      \\
                  &                       &                        & \cite{Bhanot2021ThePO}           & Mortality Prediction                                                    & MIMIC-III\cite{johnson2016mimic}                                                                            & EHR                \\
                  & Data Purification     &                        & \cite{Meng2021MIMICIFIA}         & Mortality Prediction                                                    & MIMIC-IV\cite{johnson2020mimic}                                                                             & EHR                \\
                  &                       &                        & \cite{Minot2021InterpretableBM}  & Health Condition Classification                                         & n2c2\cite{kumar2015creation}, MIMIC-III\cite{johnson2016mimic}            & EHR, Clinical Note \\
Model Development & Model Desensitization & Adversarial Learning   & \cite{Correa2021TwostepAD}       & Radiology Findings Identification                                       & Private                                                                                                                                       & Medical Image      \\
                  &                       &                        & \cite{creager2019flexibly}       & ASCVD Classification                                                    & Stanford Medicine Research Data Repository\cite{lowe2009stride}                                             & EHR                \\
                  &                       &                        & \cite{boughorbel2021fairness}    & In Hospital Mortality Prediction, Patient Membership Prediction         & MIMIC-III\cite{johnson2016mimic}                                                                           & EHR, Clinical Note \\
                  &                       &                        & \cite{zhao2020training}          & HIV Diagnosis, Morphological Sex Identification, Bone Age Determination & HIV Dataset\cite{pfefferbaum2018accelerated},NCANDA dataset\cite{tibshirani1993introduction},Bone-aging Dataset\cite{halabi2019rsna} & EHR, MRI           \\
                  &                       & Disentanglement        & \cite{boughorbel2021fairness}    & Appointment No-show Prediction                                          & Private                                                                                                                                       & EHR                \\
                  &                       & Contrastive Learning   & \cite{gorade2023pacl}            & Chest X-ray Classification                                              & NIH-ChestXRay8\cite{Wang2017ChestXRay8HC}                                                                   & Medical Image      \\
                  & Model Constraint      &                        & \cite{Pfohl2019CounterfactualRF} & Inpatient Mortality Prediction, Length of Stay Prediction               & Stanford Medicine Research Data Repository\cite{lowe2009stride}                                             & EHR                \\
Model Deployment  & Decision Explanation  &                        & \cite{Meng2021MIMICIFIA}         & In Hospital Mortality Prediction                                        & MIMIC-IV\cite{johnson2020mimic}                                                                             & EHR                \\
                  & Model Adjustment      &                        & \cite{Jabbour2020DeepLA}         & Congestive Heart Failure Prediction                                     & MIMIC-CXR\cite{Johnson2019MIMICCXRAL},~CheXpert\cite{Irvin2019CheXpertAL} & Medical Image      \\
                  & Outcome Adjustment    &                        & \cite{pfohl2022net}              & Ten-year Atherosclerotic Cardiovascular Disease (ASCVD) Risk Prediction & Optum CDM\cite{OptumHea78:online}                                                                           & EHR                
\end{tblr}
\end{table*}

In the discourse of algorithmic fairness, both shortcut learning and confounding effects epitomize pathways through which machine learning models may inadvertently perpetuate biases. Shortcut learning occurs when models exploit easy but unreliable correlations to make predictions~\cite{brown2023detecting,banerjee2023shortcuts}, often bypassing more substantive but complex relationships.
For instance, a study shows that models can capture and amplify the association between labels and sensitive attributes, even in balanced datasets~\cite{wang2019balanced}. The learned model may amplify the association between label and gender, mimicking an imbalanced dataset.
This is akin to a model using confounding variables that correlate with both the input features and the output labels, thus rendering the predictions unfair \cite{zhao2020training,deng2022genopathomic,glocker2023algorithmic}. The recent literature underscores the similarity between these phenomena: both are manifestations of models' proclivity to capitalize on spurious correlations rather than causally relevant patterns. Such practices not only compromise the equity of the models but also their robustness and reliability. Confounder-aware approaches and mitigation strategies, as delineated in seminal works, are therefore critical in ensuring that machine learning contributes to the fair and just application of AI in healthcare, and does not inadvertently exacerbate existing disparities.
{\color{blue}
Typically, machine learning models aim to maximize overall predictive performance on the training data. This focus may lead to optimizing for individuals that occur more frequently, while neglecting underrepresented groups due to sampling bias. Consequently, a model may exhibit superior overall performance but fail to generalize well for underrepresented groups~\cite{chen2018my}. For instance, in the field of radiology, convolutional neural networks (CNNs) have been found to exhibit inconsistencies in diagnosis, particularly for underserved groups such as Hispanic patients and Medicaid recipients in the United States, leading to a higher rate of underdiagnosis or misdiagnosis compared to White patients~\cite{seyyed2021underdiagnosis}.
Furthermore, studies suggest that different machine learning algorithms can exhibit varying degrees of bias when applied to the same dataset~\cite{yuan2021assessing}.
The study assessed Logistic Regression, Random Forest, and XGBoost for their performance and fairness in healthcare tasks like predicting hospital stays and diagnosing diseases. It highlighted significant variations in how these algorithms handled sensitive data like race and gender across identical datasets.
}

\subsection{Bias in Model Deployment}
A trained machine learning model can be applied to clinical practice when it has passed regulatory authorization. Bias is likely to occur at this stage,
and there may contain
two types of bias (as shown in Figure~\ref{fig:pipe}(c)).

\subsubsection{Training–serving Skew Bias}
The training service skew bias is due to the fact that the data distribution encountered by the model in the deployment environment is different from the data distribution at the time of training. This phenomenon is known as the distributional shift~\cite{Chen2021AlgorithmFI,subbaswamy2020development}.
During model training, a strong assumption is that the training and test datasets are drawn independently and exactly from the same distribution (i.i.d.).
This can lead to fairness problems when the model is deployed, even if it satisfies the notion of fairness in the training dataset.
The phenomenon of distributional shift can occur with racially skewed public biobank datasets, which has a differential impact on ethnic subpopulations. For example, the first AI model to surpass clinical rank in predicting lymph node metastasis was trained and evaluated on the CAMELYON16/17 dataset, which is unique to the Netherlands~\cite{EhteshamiBejnordi2017DiagnosticAO,Howard2021TheIO}.
In addition to changes in ethnicity in the population, changes in medical equipment, such as image capture and biometrics, can also lead to bias. For example, in radiology, there may be differences in radiation dose that affect the signal-to-noise ratio of the images obtained. In pathology, there is also a great deal of heterogeneity in tissue preparation, staining protocols, and specific scanner camera parameters, which has been shown to affect model performance in cancer diagnostic tasks~\cite{Chen2021AlgorithmFI,castro2020causality}.
Data sets may also change in response to technological developments or changes in human behavior. A typical example includes the migration of ICD-8 to ICD-9~\cite{Heslin2017TrendsIO}. Another example is the discontinuation of the Epic sepsis model (ESM) due to changes in patient demographics as a result of COVID-19~\cite{Chen2021AlgorithmFI}. 

Most of the work has focused on short-term learning of fairness classifiers, and there has been few research on the analysis of fairness metrics under temporal or spatial dataset transfer. One work~\cite{schrouffdiagnosing} uses a causal framing help diagnose failures of fairness transfer.

\subsubsection{Interaction Bias}\label{sec:inter_bias}
This type of bias arises from the interaction of the model with its users. On the one hand, protected groups may distrust a model's predictions in light of a history of exploitation and unethical behavior, believing that the model is biased against them.
This is also referred as {informed mistrust bias}~\cite{giovanola2022beyond}.
On the other hand, clinicians can also place too much trust in machine learning models and inappropriately act on inaccurate predictions, which can be called {automation bias}~\cite{Rajkomar2018EnsuringFI}.

\section{Mitigation of Fairness Problems}\label{sec:mitigate}
A variety of approaches have been developed to address fairness concerns in machine learning applications within the healthcare domain. These methods can be categorized based on the stage of the machine learning life cycle at which they are applied. We delineate these approaches across three key stages: data collection, model development, and model deployment.
Furthermore, we meticulously align the motivations behind each mitigation method with the sources of bias identified in the previous section, providing a cohesive overview of how these strategies correspond to specific biases encountered in the machine learning pipeline.
Table~\ref{fig:miti_cat} presents a taxonomy of mitigation methods utilized in the domain of fair machine learning for healthcare, detailing the specific tasks, datasets, and data types to which they are applied.


\begin{figure*}[h]
    \centering
    \includegraphics[width=1.0\linewidth]{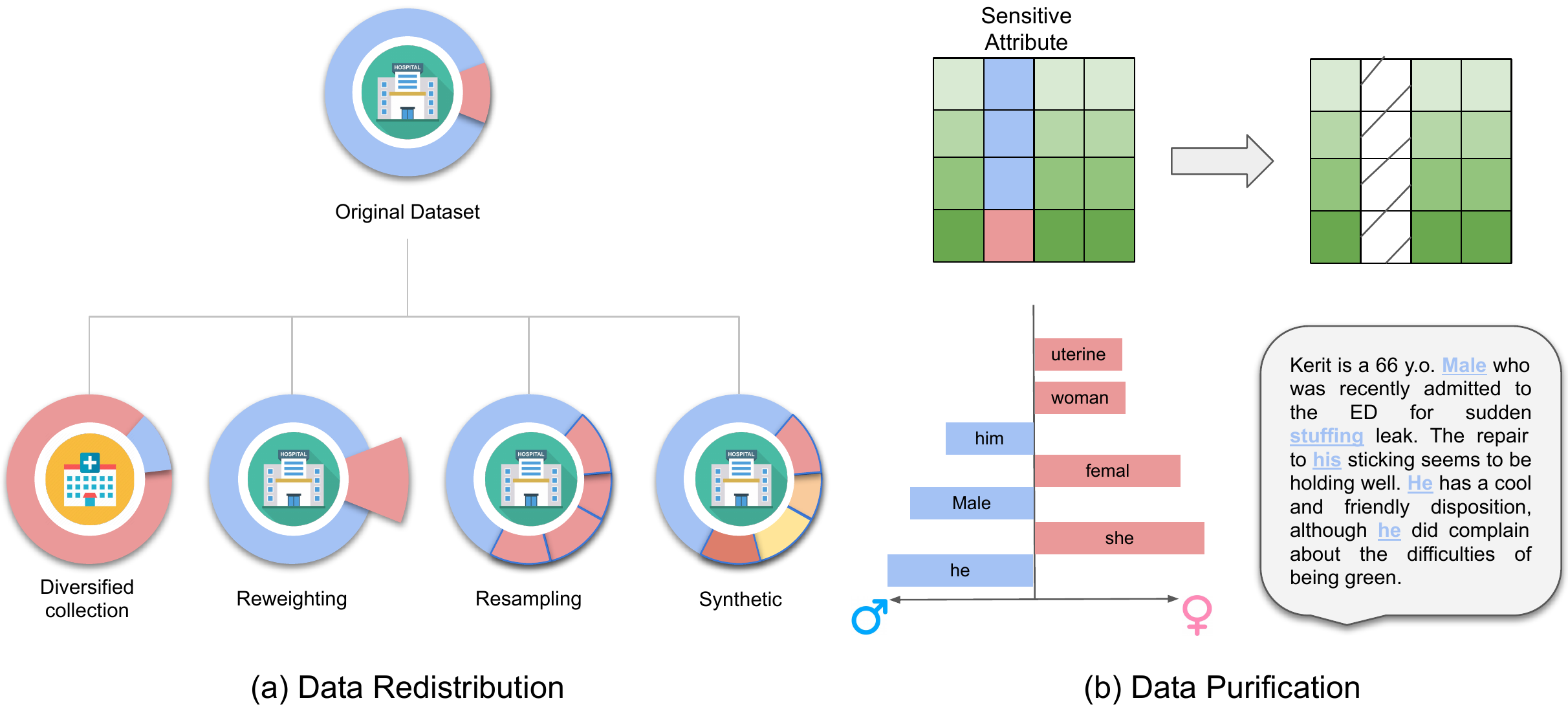}
    \vspace{-25pt}
    \caption{\textbf{An illustration of methods mitigating the fairness problem in data collection stage:} (a) Data redistribution methods adjust the distribution of the data. The diversified collection method collects data from other hospitals. The reweighting method assigns the weights to minority data. The resampling method seeks to create fair training samples in the sampling strategy. The synthetic method generates fake data. (b) Data purification methods remove sensitive information directly from the data. For example, removing sensitive attributes from tabular data or removing gender-specific pronouns from textual data.}
    \label{fig:mitigate_data}
\end{figure*} 

\subsection{Mitigating Fairness \textcolor{blue}{Problems} in Data Collection}
Data bias can be transferred and embedded in machine learning models. Therefore, we can mitigate fairness problems during the data collection phase. These methods are divided into two groups, data redistribution methods and data purification methods.

\subsubsection{Data Redistribution} 
Data distribution discrepancies, as discussed in Section~\ref{sec_bias_in_dc}, often lead to fairness problems in machine learning models. For instance, minority bias arises from the imbalanced data of different demographic groups, while missing data bias emerges due to the uneven distribution of unseen data. Several data redistribution techniques aim to rectify these imbalances, including diversified collection, data reweighting, data resampling, and data synthesis.

\paragraph{Diversified Collection}
While the direct collection of more diverse data is a straightforward solution, practical challenges like patient privacy and data collection costs often hinder such efforts. Federated learning offers a solution by enabling model training across multiple decentralized datasets without directly sharing the data~\cite{Chen2021AlgorithmFI,li2023targeting}. An instance of this is Swarm Learning (SL) which, when evaluated on the Skin ISIC 2018 dataset, exhibited enhanced fairness compared to centralized training~\cite{Fan2021OnTF}. However, federated learning does not guarantee balanced data.

\paragraph{Data Reweighting}
By assigning importance weights to training data, reweighting adjusts for data distribution imbalances. Applications of reweighting are seen in skin lesion classification and Alzheimer’s disease diagnosis~\cite{Xue2019RobustLA, Wachinger2016DomainAF}. A notable drawback is that models trained with weighted samples might lack robustness, leading to estimator variance.

\paragraph{Data Resampling}
Resampling rectifies underrepresentation by adjusting the sub-samples of the original dataset. Techniques like SMOTE combine oversampling of minority groups with undersampling of majority ones, proving beneficial in tasks like heart failure survival prediction~\cite{Chawla2002SMOTESM}. However, such methods may reduce the diversity of data characteristics.

\paragraph{Synthetic Data}
Synthetic data, often generated using algorithms like GANs, can enhance data distribution~\cite{Tiwald2021RepresentativeF, rajotte2021reducing}. By imposing fairness constraints during the generation process, biases in synthetic data can be controlled. 

Synthetic data alleviates data privacy and cost concerns~\cite{Bhanot2021ThePO}, consistent with HIPAA's stipulations~\cite{edemekong2018health}. It generates de-identified datasets that preserve statistical properties without revealing personal health information (PHI), thus supporting HIPAA's objective to protect patient privacy. Federated learning enhances this by allowing institutions to collaboratively train models while each entity maintains control over its PHI, a process in harmony with HIPAA's privacy and security rules.

\subsubsection{Data Purification}

Data purification approaches aim to mitigate fairness problems by adjusting data features or labels, often by addressing biases related to sensitive attributes.

\paragraph{Removing Sensitive Attributes}
A common intuition in data purification is to remove sensitive attributes from the dataset, a method known as fairness through unawareness. However, this approach has limitations, as protected attributes can still be inferred from other features or their combinations, which act as proxy variables correlating with protected group membership~\cite{Meng2021MIMICIFIA}.

\paragraph{Mitigation in Language Models}
In the realm of Natural Language Processing (NLP), data purification has been explored for clinical notes. One study~\cite{Minot2021InterpretableBM} quantified the ``genderedness" of n-grams in clinical notes using cosine similarity between word vectors generated by BERT-base and Clinical BERT word embeddings~\cite{Devlin2019BERTPO, Alsentzer2019PubliclyAC}. The most biased n-grams were then identified using rank perturbation dispersion (RTD) and subsequently removed from the clinical notes~\cite{dodds2020allotaxonometry}.

\paragraph{Addressing Label Bias}
Label bias, which occurs when labels in the dataset are biased, represents another challenge. Data massaging tackles this by changing the labels of some objects in the dataset~\cite{Kamiran2011DataPT}. This method's application in healthcare remains an area yet to be explored.

\subsection{Mitigating Fairness Problems in Model Development}

As discussed in Section~\ref{sec:bias_in_mdev}, algorithmic bias during the model development stage can result in machine learning models that inherit and potentially amplify biases, leading to fairness problems. Two key drivers of this bias are identified: first, shortcut learning, where models rely on sensitive information for predictions; and second, optimization processes that fail to generalize for underrepresented groups.

To address these issues, we introduce two categories of approaches to mitigate fairness problems during the model development stage: model desensitization and model constraint. Model desensitization involves techniques that reduce a model's reliance on sensitive attributes, thereby preventing it from making biased predictions based on those attributes. On the other hand, model constraint methods impose restrictions on the model training process to ensure fair treatment of all groups, especially those underrepresented in the training data.

\begin{figure*}[t]
    \centering
    \includegraphics[width=1.0\linewidth]{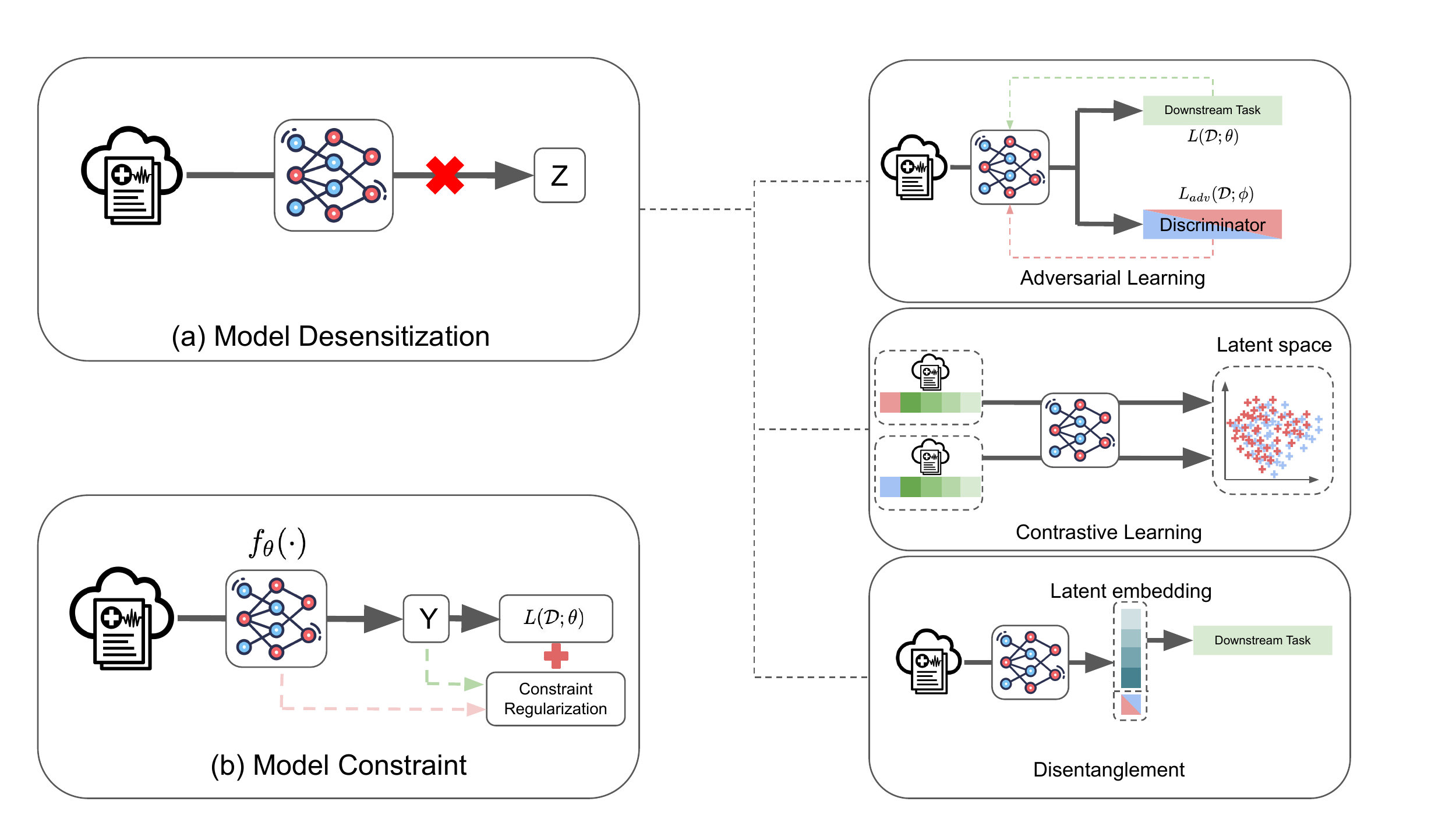}
    
    \caption{\textbf{An illustration of methods mitigating the fairness problem in model development stage}: (a) Model desensitization removes the ability of the model to discriminate between sensitive attribute information. Adversarial learning disables the model of predicting sensitive attributes. Disentanglement method separates and removes the sensitive attribute information from latent embedding. Contrastive learning enforces the samples with various sensitive attributes to be close in latent space. (b) Model constraint methods add additional constraints or regularization term.}
    \label{fig:mitigate_dev}
\end{figure*}

\subsubsection{Model Desensitization}

Model desensitization focuses on preventing models from retaining or utilizing sensitive attribute information from the data. Simply removing sensitive attributes from the data features is not a failproof solution, as machine learning models have shown the capability to differentiate sensitive information even in their absence~\cite{li2021estimating,kinyanjui2020fairness}.
To effectively mitigate fairness problems, model desensitization approaches such as adversarial learning, representation disentanglement, and contrastive learning aim to eliminate the models' ability to discriminate based on sensitive information.

Adversarial learning is a widely used method to debias a model. Specifically, the goal of adversarial learning is to allow the model to complete downstream tasks while not predicting sensitive attributes. Adversarial learning is first introduced in Generative Adversarial Networks (GANs)~\cite{goodfellow2014generative} and then applied to fair machine learning~\cite{madras2018learning}.
Adversarial learning generally contains two branches: one is for downstream tasks, while the other is to remove sensitive attribute information:
\begin{equation}
    \min_{\theta} \max_{\phi} L(\mathcal{D};\theta) + L_{adv}(\mathcal{D};\phi), \label{eqn:adversarial}
\end{equation}
where $\mathcal{D}$ is the training dataset. $\theta$ is the parameter for the downstream task and $\phi$ is the parameter for adversarial classification. $L$ is the normal object function and $L_{adv}$ is the adversarial object function that indicates the error in predicting sensitive attributes.
Adversarial learning is applied to debias a model for the diagnosis of chest X-ray and mammograms~\cite{Correa2021TwostepAD}.
The authors use CNN with two branches, where one predicts the classification target and the other predicts the sensitive attributes. The training has two steps. The first step minimizes the loss for both branches. In the second step, a flipped sign gradient of adversarial branch is backpropagated, with the aim of suppressing learning of protected variables.
Similar strategy is used to reduce the confounding effect from sensitive attribute~\cite{zhao2020training}.
In addition to the use cases for medical image data, adversarial learning has also been used to build fair machine learning models that can handle EHR data and textual data~\cite{Pfohl2019CreatingFM,Zhang2020HurtfulWQ}.

Some other model desensitization approaches have been proposed in the context of general fairness problems instead of healthcare.
The disentanglement method assumes that the entangled information from the input space could be disentangled in the latent embedding space. To make downstream tasks fair, the disentanglement method separates and removes sensitive information from the latent embedding space. Existing work has explored the use of the Variational Autoencoder (VAE) to achieve group and subgroup fairness with respect to multiple sensitive attributes~\cite{creager2019flexibly,boughorbel2021fairness}.
The contrastive learning method projects the input data into the latent space and encourages data points with various sensitive attributes to be close in the latent space and data points with the same sensitive attributes to be scattered. Some work has explored the use of contrastive learning methods to debias the pre-trained text encoder~\cite{cheng2021fairfil}, image encoder~\cite{gorade2023pacl}, or to remove the effect of gender information on self-supervised embedding~\cite{tsai2021conditional}.

\subsubsection{Model Constraint}

Addressing another potential driver of algorithmic bias—namely, the failure of the optimization goal to generalize to underrepresented groups—model constraint methods take a direct approach. Unlike model desensitization methods that implicitly debias the model, model constraint methods mitigate fairness problems by explicitly incorporating constraints into the optimization goal.

This often involves adding fairness-specific optimization objectives. For instance, these objectives might directly improve fairness metrics~\cite{Agarwal2018ARA} or include regularization terms to enforce non-discrimination principles or counterfactual fairness~\cite{Pfohl2019CounterfactualRF}. One notable approach involves developing an augmented counterfactual fairness criterion to reduce biases in Electronic Health Record (EHR) data. This method requires the machine learning model to make consistent predictions for a patient and a counterfactual version of the patient after altering the sensitive attribute. The optimization objective function comprises three components: prediction losses for factual and counterfactual samples, and an additional regularization term designed to meet the proposed fairness criteria~\cite{garg2019counterfactual}.

Despite their direct approach to addressing fairness, model constraint methods are not without drawbacks. It has been observed that stringent optimization constraints can sometimes reduce predictive performance. Moreover, the impact of regularization strength on fairness metrics can vary, presenting challenges in balancing performance and fairness.



\begin{figure*}[h]
    \centering
    \includegraphics[width=0.8\linewidth]{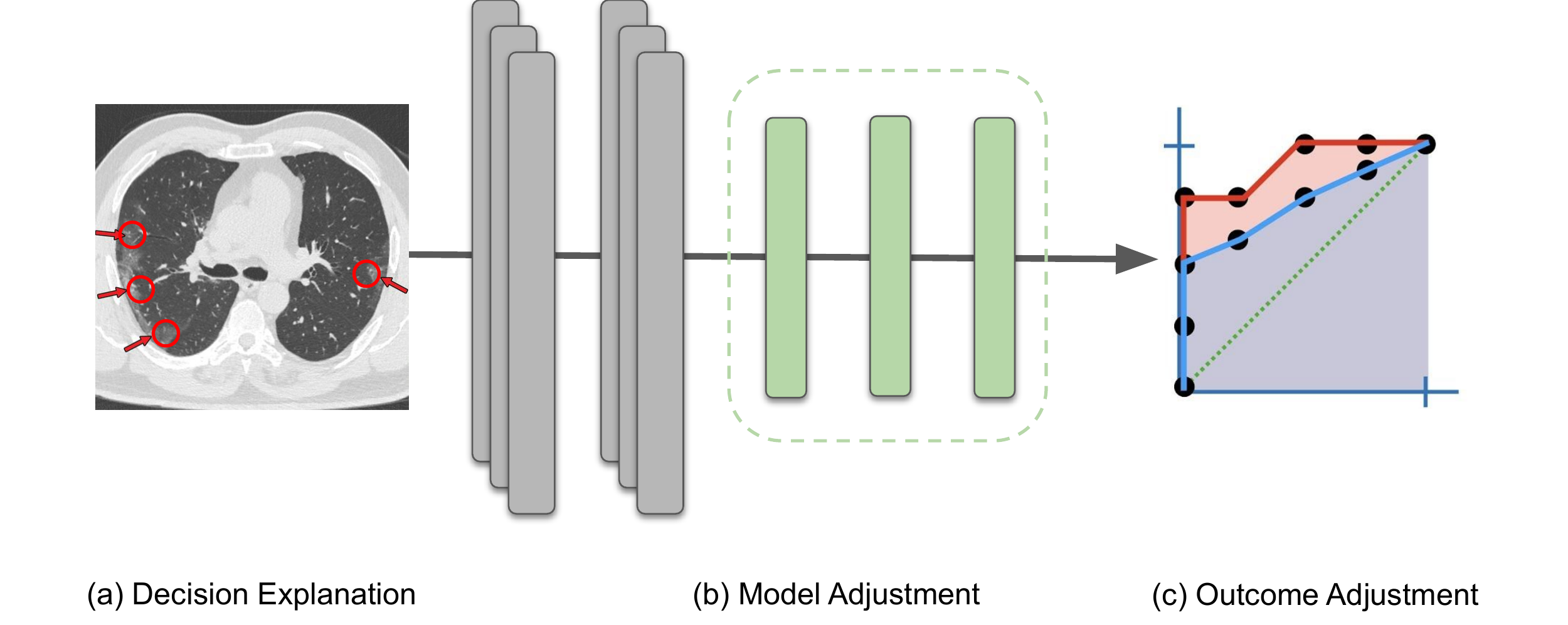}
    
    \caption{\textbf{An illustration of methods mitigating the fairness problem in model deployment stage}: (a) The decision explanation method offers the explanation to the outcome via XAI tool. (b) The model adjustment method fine-tunes the last few layers of the deployed mode. (c) The outcome adjustment method adjusts the original outcome to meet fairness requirement.}
    \label{fig:mitigate_dep}
\end{figure*}

\subsection{Mitigating Fairness Problems in Model Deployment}

Deploying machine learning models in clinical settings often surfaces biases not apparent during training or testing. Constructing entirely unbiased models from the outset is challenging and resource-intensive. Thus, post-deployment mitigation strategies are essential for addressing biases as they emerge. This section explores three key methods: decision explanation, model adjustment, and outcome adjustment, each addressing specific biases such as interaction bias and training-serving skew bias.

\subsubsection{Decision Explanation}

Fairness in deployed machine learning systems is not solely a technical challenge but a socio-technical one, where human interaction with the model is pivotal. Interaction bias, as delineated in Section~\ref{sec:inter_bias}, contributes to unfair outcomes during deployment. To mitigate such biases, the application of explainable artificial intelligence (XAI) is crucial, enabling users to understand and appropriately trust the model's decisions.

XAI can demystify model predictions, which is critical when balancing the trust in an algorithm's decisions against the risk of perpetuating unfairness. It is particularly important in healthcare, where decisions have profound implications. For instance, studies have shown that demographic features can disproportionately influence algorithmic decisions, potentially leading to differential treatment across patient groups~\cite{Meng2021MIMICIFIA}. XAI techniques have revealed such biases by highlighting the varying importance of sensitive attributes across different demographics.

Conversely, mistrust in fair models can also undermine their utility, prompting patients to eschew treatments or withhold information~\cite{Dodge2019ExplainingMA,Rajkomar2018EnsuringFI}. Addressing this, research indicates that clear explanations of model decisions can foster trust both in the system and among medical professionals~\cite{Adadi2018PeekingIT,Holzinger2019CausabilityAE}.

Ultimately, integrating fairness-oriented knowledge into XAI methods not only clarifies model decisions but also guides the refinement of models to ensure equitable outcomes~\cite{du2019LearningCD}. The synergistic relationship between fairness and explainability in machine learning models is an emergent field of research that warrants further exploration, as will be discussed in Section~\ref{challenge_ethical}.

\subsubsection{Model Adjustment}
Training-serving skew bias leads to fairness problem since it violates the assumption that data in the deployment phase are i.i.d. with the data in the training phase. The model adjustment method, such as transfer learning, seeks to solve this problem by fine-tuning part of the model. 

Since naively retraining the entire model can be expensive and impossible, transfer learning can provide a simple and effective way to mitigate the problem~\cite{Jabbour2020DeepLA}.
This work proposes solving the shortcut problem where the model relies on simple and shallow features (e.g. the sensitive attribute) to make the decision. 
Specifically, the training pipeline contains two stages. The authors first train the model on a biased dataset. Then the model is tuned on a new unbiased dataset with only the last few layers being fine-tuned. The results show that the proposed approach improves the generalization performance in older people. 

\subsubsection{Outcome Adjustment}

Outcome adjustment strategies are employed to enhance fairness for protected groups by altering the model's outputs or decision boundaries~\cite{pfohl2022net}.

Calibration, for instance, aims to align the proportion of positive predictions with the actual rate of positive outcomes across various subgroups~\cite{Dawid1982TheWB}. Fairness in this context demands that such alignment is maintained across both protected and non-protected subgroups alike~\cite{Caton2020FairnessIM}. Nevertheless, the challenge arises when calibration efforts confront the incompatibility between different fairness standards. Notably, attempts to calibrate across multiple protected groups often find themselves at odds with criteria like equalized odds or disparate impact~\cite{Pleiss2017OnFA}. This conundrum necessitates a nuanced approach to calibration, where the trade-offs between competing fairness dimensions are carefully balanced.

Thresholding takes a different tack by redefining decision boundaries. It can be particularly effective in situations where the model's default decision threshold does not accommodate the protected group adequately. By employing variable thresholds based on sensitive attributes, a model can be tuned to fulfill fairness metrics such as equal odds or equal opportunity~\cite{Hardt2016EqualityOO}. For example, applying a lower threshold for a minority group could increase their representation in positive predictions, aligning with the goal of equal opportunity.

\textbf{Discussion of the applicability of the mitigation method.}
In this review, we have previously highlighted in Section~\ref{sec:ml-health} the need for distinct forms of distributive justice in various healthcare settings, specifically equal allocation and equal performance. And it is important to note that simultaneously achieving these fairness constraints can be challenging~\cite{Mehrabi2021ASO}.

Only few of existing literature on mitigation methods study their applicability to different measures of fairness metric.
For example, the model constraint approach~\cite{Agarwal2018ARA} offers flexibility in satisfying either equal allocation or equal performance, as the fairness constraint can be incorporated as an optimization objective.
Recent studies have also demonstrated the effectiveness of certain model desensitization methods in ensuring either equal allocation or equal performance through appropriate optimization.
For instance, some work~\cite{madras2018learning} proposed the use of adversarial objects to achieve demographic parity and equalized of odds.
However, the previous two work are in the field of general fair machine learning. Few work on fair machine learning for healthcare discuss the applicability of appropriate fairness metrics, which is an obstacle to the deployment of mitigation methods in real healthcare scenarios and may even exacerbate fairness problems. As a result, we advocate for additional research efforts to conduct detailed experiments and discussions on the suitability of different fairness metrics for different mitigation methods in order to ensure their effective implementation in healthcare settings.

\section{Research Challenges}\label{sec:challenge}
Despite current progress, there are numerous research challenges that must be addressed before machine learning methods can be used in clinical practice.

\subsection{Uncertainty and Fairness in Healthcare}
Machine learning and probabilistic methods have become ubiquitous across various domains, with their application in medical data being particularly critical due to the inherent uncertainty from noise in the data. Capturing and analyzing the uncertainty in data and models is paramount, more so in high-stakes environments such as clinical settings. In such contexts, physicians might leverage the quantified uncertainty to prioritize manual review of cases that the model deems highly uncertain. The advent of new deep learning techniques has seen a significant rise in addressing such uncertainties \cite{Alizadehsani2021HandlingOU}. Despite this, the interplay between fairness and uncertainty has not been explored thoroughly in research.

Uncertainty can play a pivotal role in highlighting fairness problems within machine learning applications in healthcare \cite{Lu2021FairCP,Lu2021EvaluatingSD}. Addressing epistemic uncertainty, which arises from incomplete knowledge, often involves integrating more data into the model. On the other hand, aleatoric uncertainty, which is inherent and irreducible, demands distinct strategies. The measurement and communication of uncertainties are crucial for identifying potential unfairness in model predictions \cite{bhatt2021uncertainty}. Incorporating model uncertainty into fairness metrics can provide a more comprehensive view of model performance across different groups, ensuring that disparities in prediction confidence do not go unnoticed \cite{ali2021accounting}. An understanding of aleatoric uncertainty can lead to models that are inherently fairer, offering improved outcomes for underrepresented groups in the data \cite{tahir2023fairness}. Furthermore, active learning techniques, which focus on the selection of diverse and representative data during model training, have been proposed as a means to preemptively mitigate bias \cite{branchaud2021can}. There is a clear need for further investigation into how uncertainty impacts the fairness of machine learning models, a step that is crucial for the responsible deployment of AI in sensitive sectors.

\subsection{Long-term Fairness in Healthcare}
\textcolor{blue}{Another distributive justice called equal outcome or equal benefit is not mentioned in Section~\ref{justice}.} It refers to the assurance that protected groups have the same benefit from the deployment of machine learning models.
The gap between equal allocation and equal benefit occurs when a fair decision cannot guarantee fair benefit to patients in the future.
Most current research has focused on fairness problems in machine learning in static classification scenarios and has not examined how these decisions will affect the future~\cite{heidari2019longterm}. It is often assumed that unfairness can be improved better after imposing fairness constraints on machine learning models. However, this is not the case in healthcare settings in practice.
Even in a one-step feedback model, ordinary fairness standards generally do not promote improvement over time and may cause harm~\cite{Liu2018DelayedIO}.
The key difficulty in alleviating the long-term fairness problem is to simulate the long-term dynamics and predict the future benefit~\cite{d2020fairness}.

Another research challenge is that the healthcare system is not an isolated system.
When machine learning algorithms are embedded in clinical systems, the diagnostic decisions they make are collected and combined into new clinical data. These data then have an impact on the performance of future machine learning algorithms. This is also called a feedback loop. When bias appears in the feedback loop, it can exacerbate the bias or create new biases and further compromise the benefit of certain demographic groups.
A similar feedback loop has been discussed in the context of the recommender system~\cite{zhang2021recommendation}.
To the best of our knowledge, no research has been conducted on the long-term fairness problem in the context of the healthcare domain.
We encourage more work on the long-term fairness of machine learning algorithms in healthcare, in particular on equal benefit and feedback loop fairness in clinical applications.

\subsection{Fairness of Multi-modality Model for Healthcare}\label{challenge_multimodal}
A research question is described as multimodal when it includes multiple data types.
The human experience of the world is multimodal. Multi-modal machine learning aims to build models that can process and correlate information from multiple modalities, thus enabling advances in artificial intelligence in understanding the world around us.
One of the key driving forces of the intelligent medical system is the multimodal method. The combination of different modalities of healthcare data, each providing information about a patient's treatment from a specific perspective, overlays and complements each other to further improve the accuracy of diagnosis and treatment.
For example, the visual quest answering task~\cite{antol2015vqa} combines computer vision and natural language processing, and the model can answer relevant questions based on medical images and clinical notes~\cite{lin2021medical}.
However, multimodal models face more serious bias and fairness problems than uni-modal models, despite improvements in performance~\cite{Booth2021BiasAF}.
Only a few works have focused on multimodality fairness problems in healthcare systems~\cite{Chen2020ExploringTS}.
The forms in which bias exists vary across modality data, as do the methods used to mitigate it.
As previous work has focused on the fairness problem in uni-modal data, we encourage the discovery and mitigation of bias in healthcare of multimodal data.

\subsection{Ethical Machine Learning in Healthcare}\label{challenge_ethical}
The ethical landscape of machine learning within healthcare encompasses pivotal concepts such as fairness, interpretability, privacy, robustness, and security. These facets are deeply intertwined, with their relationships characterized by both synergy and tension.
Fairness in healthcare AI seeks to ensure equitable treatment and outcomes across diverse patient groups. Interpretability contributes to this goal by demystifying model predictions, thereby fostering trust and enabling the identification of potential biases—critical in a clinical setting~\cite{du2019techniques,Howard2021TheIO,Meng2021MIMICIFIA}. However, the pursuit of fairness may inadvertently conflict with privacy, particularly for underprivileged groups who may suffer disproportionate privacy losses~\cite{chang2021privacy}.
Conversely, the alliance between robustness and fairness is more harmonious in healthcare AI. Robust fair training aims to inoculate models against perturbations that could skew decision-making, thus safeguarding equitable outcomes~\cite{lee2021machine}. This is paramount in clinical environments where decisions must remain stable despite data variability and adversarial conditions.
Furthermore, the convergence of differential privacy and adversarial robustness underscores a promising avenue where privacy-preserving techniques also fortify models against malicious attacks, a duality of particular relevance to safeguarding sensitive health data~\cite{pinot2019unified}.
Yet, the interplay of interpretability, fairness, robustness, and privacy in healthcare AI is nascent. Research often probes these dimensions in isolation, seldom navigating their intersections. Given their mutual reinforcement and constraints, an integrated approach is imperative. Advancing multi-faceted ethical frameworks that concurrently address these dimensions will be instrumental in realizing the full potential of AI in healthcare—delivering models that are not only technically proficient but also ethically sound and clinically viable.

\section{Conclusions}
In this survey, we have synthesized the existing literature on the intersection of machine learning and fairness within healthcare. 
\textcolor{blue}{
Drawing from the foundational work in distributive justice, we have applied the classification of fairness problems in healthcare-focused machine learning methods, as identified by existing research, into two principal categories: equal allocation and equal performance.}
This has allowed us to map the metrics commonly used in fair machine learning to these categories specifically in the healthcare context.
We have delineated biases according to the three distinct stages of the machine learning lifecycle: data collection, model development, and model deployment. For each stage, we have discussed targeted mitigation methods and examined their interconnections with the sources of bias they aim to address.
Our survey reveals a gap in the critical evaluation of the effectiveness of these mitigation methods when applied to healthcare-specific fairness metrics.
We underscore the pressing nature of fairness concerns in healthcare machine learning applications and propose future research directions that promise to address these challenges.

\section{Acknowledgement}
We extend our sincere thanks for the support from the National Institutes of Health (NIH) grant 1OT2OD032581-02-211 and the National Science Foundation (NSF) grants IIS 1900990, 1939716, and 2239257, which have significantly contributed to this survey paper.
\bibliographystyle{abbrv}
\bibliography{ref}

\begin{thebibliography}{100}

\bibitem{AzureHea70:online}
Azure health bot | microsoft azure.
\newblock
  \url{https://azure.microsoft.com/en-us/products/bot-services/health-bot}.
\newblock (Accessed on 01/02/2024).

\bibitem{ALargeLa55:online}
A large language model for healthcare | nhs-llm and opengpt.
\newblock
  \url{https://aiforhealthcare.substack.com/p/a-large-language-model-for-healthcare}.
\newblock (Accessed on 01/02/2024).

\bibitem{OptumHea78:online}
Optum - health services innovation company.
\newblock \url{https://www.optum.com/}.
\newblock (Accessed on 11/07/2023).

\bibitem{Adadi2018PeekingIT}
A.~Adadi and M.~Berrada.
\newblock Peeking inside the black-box: A survey on explainable artificial
  intelligence (xai).
\newblock {\em IEEE Access}, 6:52138--52160, 2018.

\bibitem{adamson2018machine}
A.~S. Adamson and A.~Smith.
\newblock Machine learning and health care disparities in dermatology.
\newblock {\em JAMA dermatology}, 154(11):1247--1248, 2018.

\bibitem{Agarwal2018ARA}
A.~Agarwal, A.~Beygelzimer, M.~Dud{\'i}k, J.~Langford, and H.~M. Wallach.
\newblock A reductions approach to fair classification.
\newblock {\em ArXiv}, abs/1803.02453, 2018.

\bibitem{ahmed2021examining}
S.~Ahmed, C.~T. Nutt, N.~D. Eneanya, P.~P. Reese, K.~Sivashanker, M.~Morse,
  T.~Sequist, and M.~L. Mendu.
\newblock Examining the potential impact of race multiplier utilization in
  estimated glomerular filtration rate calculation on african-american care
  outcomes.
\newblock {\em Journal of general internal medicine}, 36(2):464--471, 2021.

\bibitem{ali2021accounting}
J.~Ali, P.~Lahoti, and K.~P. Gummadi.
\newblock Accounting for model uncertainty in algorithmic discrimination.
\newblock In {\em Proceedings of the 2021 AAAI/ACM Conference on AI, Ethics,
  and Society}, pages 336--345, 2021.

\bibitem{Alizadehsani2021HandlingOU}
R.~Alizadehsani, M.~Roshanzamir, S.~Hussain, A.~Khosravi, A.~Koohestani, M.~H.
  Zangooei, M.~Abdar, A.~Beykikhoshk, A.~Shoeibi, A.~Zare, M.~Panahiazar,
  S.~Nahavandi, D.~Srinivasan, A.~F. Atiya, and U.~R. Acharya.
\newblock Handling of uncertainty in medical data using machine learning and
  probability theory techniques: a review of 30 years (1991–2020).
\newblock {\em Annals of Operations Research}, pages 1 -- 42, 2021.

\bibitem{Alsentzer2019PubliclyAC}
E.~Alsentzer, J.~R. Murphy, W.~Boag, W.-H. Weng, D.~Jin, T.~Naumann, and
  M.~B.~A. McDermott.
\newblock Publicly available clinical bert embeddings.
\newblock {\em ArXiv}, abs/1904.03323, 2019.

\bibitem{antol2015vqa}
S.~Antol, A.~Agrawal, J.~Lu, M.~Mitchell, D.~Batra, C.~L. Zitnick, and
  D.~Parikh.
\newblock Vqa: Visual question answering.
\newblock In {\em Proceedings of the IEEE international conference on computer
  vision}, pages 2425--2433, 2015.

\bibitem{banerjee2023shortcuts}
I.~Banerjee, K.~Bhattacharjee, J.~L. Burns, H.~Trivedi, S.~Purkayastha,
  L.~Seyyed-Kalantari, B.~N. Patel, R.~Shiradkar, and J.~Gichoya.
\newblock “shortcuts” causing bias in radiology artificial intelligence:
  causes, evaluation and mitigation.
\newblock {\em Journal of the American College of Radiology}, 2023.

\bibitem{EhteshamiBejnordi2017DiagnosticAO}
B.~E. Bejnordi, M.~Veta, P.~J. van Diest, B.~van Ginneken, N.~Karssemeijer,
  G.~J.~S. Litjens, J.~A. van~der Laak, M.~Hermsen, Q.~F. Manson, M.~C.~A.
  Balkenhol, O.~G.~F. Geessink, N.~Stathonikos, M.~C. van Dijk, P.~Bult,
  F.~Beca, A.~H. Beck, D.~Wang, A.~Khosla, R.~Gargeya, H.~Irshad, A.~Zhong,
  Q.~Dou, Q.~Li, H.~Chen, H.~Lin, P.-A. Heng, C.~Hass, E.~Bruni, Q.~K.-S. Wong,
  U.~Halici, M.~{\"U}. {\"O}ner, R.~Cetin-Atalay, M.~Berseth, V.~Khvatkov,
  A.~Vylegzhanin, O.~Z. Kraus, M.~Shaban, N.~M. Rajpoot, R.~Awan,
  K.~Sirinukunwattana, T.~Qaiser, Y.-W. Tsang, D.~Tellez, J.~Annuscheit,
  P.~Hufnagl, M.~Valkonen, K.~Kartasalo, L.~Latonen, P.~Ruusuvuori,
  K.~Liimatainen, S.~Albarqouni, B.~Mungal, A.~A. George, S.~Demirci, N.~Navab,
  S.~Watanabe, S.~Seno, Y.~Takenaka, H.~Matsuda, H.~A. Phoulady, V.~A. Kovalev,
  A.~Kalinovsky, V.~Liauchuk, G.~Bueno, M.~del Milagro Fern{\'a}ndez-Carrobles,
  I.~Serrano, O.~Deniz, D.~Racoceanu, and R.~Ven{\^a}ncio.
\newblock Diagnostic assessment of deep learning algorithms for detection of
  lymph node metastases in women with breast cancer.
\newblock {\em JAMA}, 318:2199–2210, 2017.

\bibitem{Berk2018FairnessIC}
R.~A. Berk, H.~Heidari, S.~Jabbari, M.~Kearns, and A.~Roth.
\newblock Fairness in criminal justice risk assessments: The state of the art.
\newblock {\em Sociological Methods \& Research}, 50:3 -- 44, 2018.

\bibitem{Bhanot2021ThePO}
K.~Bhanot, M.~Qi, J.~S. Erickson, I.~Guyon, and K.~P. Bennett.
\newblock The problem of fairness in synthetic healthcare data.
\newblock {\em Entropy}, 23, 2021.

\bibitem{bhatt2021uncertainty}
U.~Bhatt, J.~Antor{\'a}n, Y.~Zhang, Q.~V. Liao, P.~Sattigeri, R.~Fogliato,
  G.~Melan{\c{c}}on, R.~Krishnan, J.~Stanley, O.~Tickoo, et~al.
\newblock Uncertainty as a form of transparency: Measuring, communicating, and
  using uncertainty.
\newblock In {\em Proceedings of the 2021 AAAI/ACM Conference on AI, Ethics,
  and Society}, pages 401--413, 2021.

\bibitem{Booth2021BiasAF}
B.~M. Booth, L.~Hickman, S.~K. Subburaj, L.~Tay, S.~E. Woo, and S.~K.
  D’Mello.
\newblock Bias and fairness in multimodal machine learning: A case study of
  automated video interviews.
\newblock {\em Proceedings of the 2021 International Conference on Multimodal
  Interaction}, 2021.

\bibitem{boughorbel2021fairness}
S.~Boughorbel, F.~Jarray, and A.~Kadri.
\newblock Fairness in tabnet model by disentangled representation for the
  prediction of hospital no-show.
\newblock {\em arXiv preprint arXiv:2103.04048}, 2021.

\bibitem{branchaud2021can}
F.~Branchaud-Charron, P.~Atighehchian, P.~Rodr{\'\i}guez, G.~Abuhamad, and
  A.~Lacoste.
\newblock Can active learning preemptively mitigate fairness issues?
\newblock {\em arXiv preprint arXiv:2104.06879}, 2021.

\bibitem{brogan2021next}
J.~Brogan.
\newblock The next era of biomedical research: Prioritizing health equity in
  the age of digital medicine.
\newblock {\em Voices in Bioethics}, 7, 2021.

\bibitem{brown2023detecting}
A.~Brown, N.~Tomasev, J.~Freyberg, Y.~Liu, A.~Karthikesalingam, and
  J.~Schrouff.
\newblock Detecting shortcut learning for fair medical ai using shortcut
  testing.
\newblock {\em Nature Communications}, 14(1):4314, 2023.

\bibitem{Brown2020LanguageMA}
T.~B. Brown, B.~Mann, N.~Ryder, M.~Subbiah, J.~Kaplan, P.~Dhariwal,
  A.~Neelakantan, P.~Shyam, G.~Sastry, A.~Askell, S.~Agarwal, A.~Herbert-Voss,
  G.~Krueger, T.~J. Henighan, R.~Child, A.~Ramesh, D.~M. Ziegler, J.~Wu,
  C.~Winter, C.~Hesse, M.~Chen, E.~Sigler, M.~Litwin, S.~Gray, B.~Chess,
  J.~Clark, C.~Berner, S.~McCandlish, A.~Radford, I.~Sutskever, and D.~Amodei.
\newblock Language models are few-shot learners.
\newblock {\em ArXiv}, abs/2005.14165, 2020.

\bibitem{Campanella2019ClinicalgradeCP}
G.~Campanella, M.~G. Hanna, L.~Geneslaw, A.~P. Miraflor, V.~W.~K. Silva, K.~J.
  Busam, E.~Brogi, V.~E. Reuter, D.~S. Klimstra, and T.~J. Fuchs.
\newblock Clinical-grade computational pathology using weakly supervised deep
  learning on whole slide images.
\newblock {\em Nature Medicine}, pages 1--9, 2019.

\bibitem{carbonell1983overview}
J.~G. Carbonell, R.~S. Michalski, and T.~M. Mitchell.
\newblock An overview of machine learning.
\newblock {\em Machine learning}, pages 3--23, 1983.

\bibitem{cascella2023evaluating}
M.~Cascella, J.~Montomoli, V.~Bellini, and E.~Bignami.
\newblock Evaluating the feasibility of chatgpt in healthcare: an analysis of
  multiple clinical and research scenarios.
\newblock {\em Journal of Medical Systems}, 47(1):33, 2023.

\bibitem{castro2020causality}
D.~C. Castro, I.~Walker, and B.~Glocker.
\newblock Causality matters in medical imaging.
\newblock {\em Nature Communications}, 11(1):3673, 2020.

\bibitem{Caton2020FairnessIM}
S.~Caton and C.~Haas.
\newblock Fairness in machine learning: A survey.
\newblock {\em ArXiv}, abs/2010.04053, 2020.

\bibitem{chang2021privacy}
H.~Chang and R.~Shokri.
\newblock On the privacy risks of algorithmic fairness.
\newblock In {\em 2021 IEEE European Symposium on Security and Privacy
  (EuroS\&P)}, pages 292--303. IEEE, 2021.

\bibitem{Chawla2002SMOTESM}
N.~Chawla, K.~Bowyer, L.~O. Hall, and W.~P. Kegelmeyer.
\newblock Smote: Synthetic minority over-sampling technique.
\newblock {\em J. Artif. Intell. Res.}, 16:321--357, 2002.

\bibitem{chen2018my}
I.~Chen, F.~D. Johansson, and D.~Sontag.
\newblock Why is my classifier discriminatory?
\newblock {\em Advances in neural information processing systems}, 31, 2018.

\bibitem{chen2021ethical}
I.~Y. Chen, E.~Pierson, S.~Rose, S.~Joshi, K.~Ferryman, and M.~Ghassemi.
\newblock Ethical machine learning in healthcare.
\newblock {\em Annual review of biomedical data science}, 4:123--144, 2021.

\bibitem{Chen2020ExploringTS}
J.~Chen, I.~Berlot-Attwell, S.~Hossain, X.~Wang, and F.~Rudzicz.
\newblock Exploring text specific and blackbox fairness algorithms in
  multimodal clinical nlp.
\newblock {\em ArXiv}, abs/2011.09625, 2020.

\bibitem{Chen2021AlgorithmFI}
R.~J. Chen, T.~Y. Chen, J.~Lipkov{\'a}, J.~J. Wang, D.~F.~K. Williamson, M.~Y.
  Lu, S.~Sahai, and F.~Mahmood.
\newblock Algorithm fairness in ai for medicine and healthcare.
\newblock {\em ArXiv}, abs/2110.00603, 2021.

\bibitem{cheng2021fairfil}
P.~Cheng, W.~Hao, S.~Yuan, S.~Si, and L.~Carin.
\newblock Fairfil: Contrastive neural debiasing method for pretrained text
  encoders, 2021.

\bibitem{choi2016learning}
Y.~Choi, C.~Y.-I. Chiu, and D.~Sontag.
\newblock Learning low-dimensional representations of medical concepts.
\newblock {\em AMIA Summits on Translational Science Proceedings}, 2016:41,
  2016.

\bibitem{chouldechova2017fair}
A.~Chouldechova.
\newblock Fair prediction with disparate impact: A study of bias in recidivism
  prediction instruments.
\newblock {\em Big data}, 5(2):153--163, 2017.

\bibitem{Chzhen2020FairRW}
E.~Chzhen, C.~Denis, M.~Hebiri, L.~Oneto, and M.~Pontil.
\newblock Fair regression with wasserstein barycenters.
\newblock {\em ArXiv}, abs/2006.07286, 2020.

\bibitem{codella2018skin}
N.~C. Codella, D.~Gutman, M.~E. Celebi, B.~Helba, M.~A. Marchetti, S.~W. Dusza,
  A.~Kalloo, K.~Liopyris, N.~Mishra, H.~Kittler, et~al.
\newblock Skin lesion analysis toward melanoma detection: A challenge at the
  2017 international symposium on biomedical imaging (isbi), hosted by the
  international skin imaging collaboration (isic).
\newblock In {\em 2018 IEEE 15th international symposium on biomedical imaging
  (ISBI 2018)}, pages 168--172. IEEE, 2018.

\bibitem{Correa2021TwostepAD}
R.~Correa, J.~J. Jeong, B.~Patel, H.~Trivedi, J.~W. Gichoya, and I.~Banerjee.
\newblock Two-step adversarial debiasing with partial learning - medical image
  case-studies.
\newblock {\em ArXiv}, abs/2111.08711, 2021.

\bibitem{creager2019flexibly}
E.~Creager, D.~Madras, J.-H. Jacobsen, M.~Weis, K.~Swersky, T.~Pitassi, and
  R.~Zemel.
\newblock Flexibly fair representation learning by disentanglement.
\newblock In {\em International conference on machine learning}, pages
  1436--1445. PMLR, 2019.

\bibitem{d2020fairness}
A.~D'Amour, H.~Srinivasan, J.~Atwood, P.~Baljekar, D.~Sculley, and Y.~Halpern.
\newblock Fairness is not static: deeper understanding of long term fairness
  via simulation studies.
\newblock In {\em Proceedings of the 2020 Conference on Fairness,
  Accountability, and Transparency}, pages 525--534, 2020.

\bibitem{Dawid1982TheWB}
A.~P. Dawid.
\newblock The well-calibrated bayesian.
\newblock {\em Journal of the American Statistical Association}, 77:605--610,
  1982.

\bibitem{deng2022genopathomic}
J.~Deng, J.~Yang, L.~Hou, J.~Wu, Y.~He, M.~Zhao, B.~Ni, D.~Wei, H.~Pfister,
  C.~Zhou, et~al.
\newblock Genopathomic profiling identifies signatures for immunotherapy
  response of lung adenocarcinoma via confounder-aware representation learning.
\newblock {\em Iscience}, 25(11), 2022.

\bibitem{Denis2021FairnessGI}
C.~Denis, R.~Elie, M.~Hebiri, and F.~Hu.
\newblock Fairness guarantee in multi-class classification.
\newblock 2021.

\bibitem{Devlin2019BERTPO}
J.~Devlin, M.-W. Chang, K.~Lee, and K.~Toutanova.
\newblock Bert: Pre-training of deep bidirectional transformers for language
  understanding.
\newblock In {\em NAACL}, 2019.

\bibitem{dieterich2016compas}
W.~Dieterich, C.~Mendoza, and T.~Brennan.
\newblock Compas risk scales: Demonstrating accuracy equity and predictive
  parity.
\newblock {\em Northpointe Inc}, 7(4), 2016.

\bibitem{dodds2020allotaxonometry}
P.~S. Dodds, J.~R. Minot, M.~V. Arnold, T.~Alshaabi, J.~L. Adams, D.~R.
  Dewhurst, T.~J. Gray, M.~R. Frank, A.~J. Reagan, and C.~M. Danforth.
\newblock Allotaxonometry and rank-turbulence divergence: a universal
  instrument for comparing complex systems.
\newblock {\em arXiv preprint arXiv:2002.09770}, 2020.

\bibitem{Dodge2019ExplainingMA}
J.~Dodge, Q.~V. Liao, Y.~Zhang, R.~K.~E. Bellamy, and C.~Dugan.
\newblock Explaining models: an empirical study of how explanations impact
  fairness judgment.
\newblock {\em Proceedings of the 24th International Conference on Intelligent
  User Interfaces}, 2019.

\bibitem{du2019techniques}
M.~Du, N.~Liu, and X.~Hu.
\newblock Techniques for interpretable machine learning.
\newblock {\em Communications of the ACM}, 63(1):68--77, 2019.

\bibitem{du2019LearningCD}
M.~Du, N.~Liu, F.~Yang, and X.~Hu.
\newblock Learning credible deep neural networks with rationale regularization.
\newblock {\em 2019 IEEE International Conference on Data Mining (ICDM)}, pages
  150--159, 2019.

\bibitem{Dwork2012FairnessTA}
C.~Dwork, M.~Hardt, T.~Pitassi, O.~Reingold, and R.~S. Zemel.
\newblock Fairness through awareness.
\newblock {\em ArXiv}, abs/1104.3913, 2012.

\bibitem{edemekong2018health}
P.~F. Edemekong, P.~Annamaraju, and M.~J. Haydel.
\newblock Health insurance portability and accountability act.
\newblock 2018.

\bibitem{enayati2020optimal}
S.~Enayati and O.~Y. {\"O}zalt{\i}n.
\newblock Optimal influenza vaccine distribution with equity.
\newblock {\em European Journal of Operational Research}, 283(2):714--725,
  2020.

\bibitem{Escobar2016PilotingEM}
G.~J. Escobar, B.~J. Turk, A.~I. Ragins, J.~Ha, B.~Hoberman, S.~M. Levine,
  M.~A. Ballesca, V.~X. Liu, and P.~Kipnis.
\newblock Piloting electronic medical record-based early detection of inpatient
  deterioration in community hospitals.
\newblock {\em Journal of hospital medicine}, 11 Suppl 1:S18--S24, 2016.

\bibitem{fabris2023measuring}
A.~Fabris, A.~Esuli, A.~Moreo, and F.~Sebastiani.
\newblock Measuring fairness under unawareness of sensitive attributes: A
  quantification-based approach.
\newblock {\em Journal of Artificial Intelligence Research}, 76:1117--1180,
  2023.

\bibitem{Fan2021OnTF}
D.~Fan, Y.~Wu, and X.~Li.
\newblock On the fairness of swarm learning in skin lesion classification.
\newblock {\em ArXiv}, abs/2109.12176, 2021.

\bibitem{fletcher2021addressing}
R.~R. Fletcher, A.~Nakeshimana, and O.~Olubeko.
\newblock Addressing fairness, bias, and appropriate use of artificial
  intelligence and machine learning in global health, 2021.

\bibitem{friedler2016possibility}
S.~A. Friedler, C.~Scheidegger, and S.~Venkatasubramanian.
\newblock On the (im) possibility of fairness.
\newblock {\em arXiv preprint arXiv:1609.07236}, 2016.

\bibitem{Gao2013IntegrativeAO}
J.~Gao, B.~A. Aksoy, U.~Dogrusoz, G.~Dresdner, B.~E. Gross, S.~O. Sumer,
  Y.~Sun, A.~S. Jacobsen, R.~Sinha, E.~Larsson, E.~G. Cerami, C.~Sander, and
  N.~D. Schultz.
\newblock Integrative analysis of complex cancer genomics and clinical profiles
  using the cbioportal.
\newblock {\em Science Signaling}, 6:pl1 -- pl1, 2013.

\bibitem{garg2019counterfactual}
S.~Garg, V.~Perot, N.~Limtiaco, A.~Taly, E.~H. Chi, and A.~Beutel.
\newblock Counterfactual fairness in text classification through robustness,
  2019.

\bibitem{giovanola2022beyond}
B.~Giovanola and S.~Tiribelli.
\newblock Beyond bias and discrimination: redefining the ai ethics principle of
  fairness in healthcare machine-learning algorithms.
\newblock {\em AI \& society}, pages 1--15, 2022.

\bibitem{glocker2023algorithmic}
B.~Glocker, C.~Jones, M.~Bernhardt, and S.~Winzeck.
\newblock Algorithmic encoding of protected characteristics in chest x-ray
  disease detection models.
\newblock {\em Ebiomedicine}, 89, 2023.

\bibitem{goodfellow2014generative}
I.~J. Goodfellow, J.~Pouget-Abadie, M.~Mirza, B.~Xu, D.~Warde-Farley, S.~Ozair,
  A.~Courville, and Y.~Bengio.
\newblock Generative adversarial networks, 2014.

\bibitem{gorade2023pacl}
V.~Gorade, S.~Mittal, and R.~Singhal.
\newblock Pacl: Patient-aware contrastive learning through metadata refinement
  for generalized early disease diagnosis.
\newblock {\em Computers in Biology and Medicine}, page 107569, 2023.

\bibitem{Gutman2018SkinLA}
D.~A. Gutman, N.~C.~F. Codella, M.~E. Celebi, B.~Helba, M.~A. Marchetti, N.~K.
  Mishra, and A.~C. Halpern.
\newblock Skin lesion analysis toward melanoma detection: A challenge at the
  2017 international symposium on biomedical imaging (isbi), hosted by the
  international skin imaging collaboration (isic).
\newblock {\em 2018 IEEE 15th International Symposium on Biomedical Imaging
  (ISBI 2018)}, pages 168--172, 2018.

\bibitem{halabi2019rsna}
S.~S. Halabi, L.~M. Prevedello, J.~Kalpathy-Cramer, A.~B. Mamonov, A.~Bilbily,
  M.~Cicero, I.~Pan, L.~A. Pereira, R.~T. Sousa, N.~Abdala, et~al.
\newblock The rsna pediatric bone age machine learning challenge.
\newblock {\em Radiology}, 290(2):498--503, 2019.

\bibitem{hardt2016equality}
M.~Hardt, E.~Price, and N.~Srebro.
\newblock Equality of opportunity in supervised learning.
\newblock {\em Advances in neural information processing systems},
  29:3315--3323, 2016.

\bibitem{Hardt2016EqualityOO}
M.~Hardt, E.~Price, and N.~Srebro.
\newblock Equality of opportunity in supervised learning.
\newblock In {\em NIPS}, 2016.

\bibitem{heidari2019longterm}
H.~Heidari, V.~Nanda, and K.~P. Gummadi.
\newblock On the long-term impact of algorithmic decision policies: Effort
  unfairness and feature segregation through social learning, 2019.

\bibitem{Heslin2017TrendsIO}
K.~C. Heslin, P.~L. Owens, Z.~Karaca, M.~L. Barrett, B.~J. Moore, and
  A.~Elixhauser.
\newblock Trends in opioid-related inpatient stays shifted after the us
  transitioned to icd-10-cm diagnosis coding in 2015.
\newblock {\em Medical Care}, 55:918–923, 2017.

\bibitem{Holzinger2019CausabilityAE}
A.~Holzinger, G.~Langs, H.~Denk, K.~Zatloukal, and H.~M{\"u}ller.
\newblock Causability and explainability of artificial intelligence in
  medicine.
\newblock {\em Wiley Interdisciplinary Reviews. Data Mining and Knowledge
  Discovery}, 9, 2019.

\bibitem{Howard2021TheIO}
F.~M. Howard, J.~M. Dolezal, S.~E. Kochanny, J.~J. Schulte, H.~I.-H. Chen,
  L.~R. Heij, D.~Huo, R.~Nanda, O.~I. Olopade, J.~N. Kather, N.~A. Cipriani,
  R.~L. Grossman, and A.~T. Pearson.
\newblock The impact of site-specific digital histology signatures on deep
  learning model accuracy and bias.
\newblock {\em Nature Communications}, 12, 2021.

\bibitem{Irvin2019CheXpertAL}
J.~A. Irvin, P.~Rajpurkar, M.~Ko, Y.~Yu, S.~Ciurea-Ilcus, C.~Chute,
  H.~Marklund, B.~Haghgoo, R.~L. Ball, K.~S. Shpanskaya, J.~Seekins, D.~A.
  Mong, S.~S. Halabi, J.~K. Sandberg, R.~Jones, D.~B. Larson, C.~Langlotz,
  B.~N. Patel, M.~P. Lungren, and A.~Ng.
\newblock Chexpert: A large chest radiograph dataset with uncertainty labels
  and expert comparison.
\newblock In {\em AAAI}, 2019.

\bibitem{Jabbour2020DeepLA}
S.~Jabbour, D.~F. Fouhey, E.~A. Kazerooni, M.~W. Sjoding, and J.~Wiens.
\newblock Deep learning applied to chest x-rays: Exploiting and preventing
  shortcuts.
\newblock In {\em MLHC}, 2020.

\bibitem{jiang2021generalized}
Z.~Jiang, X.~Han, C.~Fan, F.~Yang, A.~Mostafavi, and X.~Hu.
\newblock Generalized demographic parity for group fairness.
\newblock In {\em International Conference on Learning Representations}, 2021.

\bibitem{johnson2020mimic}
A.~Johnson, L.~Bulgarelli, T.~Pollard, S.~Horng, L.~Celi, and R.~Mark.
\newblock Mimic-iv (version 0.4), physionet, 2020.

\bibitem{johnson2016mimic}
A.~E. Johnson, T.~J. Pollard, L.~Shen, H.~L. Li-Wei, M.~Feng, M.~Ghassemi,
  B.~Moody, P.~Szolovits, L.~A. Celi, and R.~G. Mark.
\newblock Mimic-iii, a freely accessible critical care database.
\newblock {\em Scientific data}, 3(1):1--9, 2016.

\bibitem{Johnson2019MIMICCXRAL}
A.~E.~W. Johnson, T.~J. Pollard, S.~J. Berkowitz, N.~R. Greenbaum, M.~P.
  Lungren, C.~ying Deng, R.~G. Mark, and S.~Horng.
\newblock Mimic-cxr: A large publicly available database of labeled chest
  radiographs.
\newblock {\em ArXiv}, abs/1901.07042, 2019.

\bibitem{Kamiran2011DataPT}
F.~Kamiran and T.~Calders.
\newblock Data preprocessing techniques for classification without
  discrimination.
\newblock {\em Knowledge and Information Systems}, 33:1--33, 2011.

\bibitem{kinyanjui2020fairness}
N.~M. Kinyanjui, T.~Odonga, C.~Cintas, N.~C. Codella, R.~Panda, P.~Sattigeri,
  and K.~R. Varshney.
\newblock Fairness of classifiers across skin tones in dermatology.
\newblock In {\em International Conference on Medical Image Computing and
  Computer-Assisted Intervention}, pages 320--329. Springer, 2020.

\bibitem{kumar2015creation}
V.~Kumar, A.~Stubbs, S.~Shaw, and {\"O}.~Uzuner.
\newblock Creation of a new longitudinal corpus of clinical narratives.
\newblock {\em Journal of biomedical informatics}, 58:S6--S10, 2015.

\bibitem{kuppler2021distributive}
M.~Kuppler, C.~Kern, R.~L. Bach, and F.~Kreuter.
\newblock Distributive justice and fairness metrics in automated
  decision-making: How much overlap is there?
\newblock {\em arXiv preprint arXiv:2105.01441}, 2021.

\bibitem{Kusner2017CounterfactualF}
M.~J. Kusner, J.~R. Loftus, C.~Russell, and R.~Silva.
\newblock Counterfactual fairness.
\newblock In {\em NIPS}, 2017.

\bibitem{Kwak2019DeepHealthRA}
G.~H. Kwak and P.~Hui.
\newblock Deephealth: Review and challenges of artificial intelligence in
  health informatics.
\newblock {\em arXiv: Learning}, 2019.

\bibitem{lamont2017distributive}
J.~Lamont.
\newblock {\em Distributive justice}.
\newblock Routledge, 2017.

\bibitem{Larrazabal2020GenderII}
A.~J. Larrazabal, N.~Nieto, V.~Peterson, D.~H. Milone, and E.~Ferrante.
\newblock Gender imbalance in medical imaging datasets produces biased
  classifiers for computer-aided diagnosis.
\newblock {\em Proceedings of the National Academy of Sciences of the United
  States of America}, 117:12592 -- 12594, 2020.

\bibitem{lee2021machine}
J.-G. Lee, Y.~Roh, H.~Song, and S.~E. Whang.
\newblock Machine learning robustness, fairness, and their convergence.
\newblock In {\em Proceedings of the 27th ACM SIGKDD Conference on Knowledge
  Discovery \& Data Mining}, pages 4046--4047, 2021.

\bibitem{li2023targeting}
S.~Li, T.~Cai, and R.~Duan.
\newblock Targeting underrepresented populations in precision medicine: A
  federated transfer learning approach.
\newblock {\em The Annals of Applied Statistics}, 17(4):2970--2992, 2023.

\bibitem{li2021estimating}
X.~Li, Z.~Cui, Y.~Wu, L.~Gu, and T.~Harada.
\newblock Estimating and improving fairness with adversarial learning.
\newblock {\em arXiv preprint arXiv:2103.04243}, 2021.

\bibitem{lin2021medical}
Z.~Lin, D.~Zhang, Q.~Tac, D.~Shi, G.~Haffari, Q.~Wu, M.~He, and Z.~Ge.
\newblock Medical visual question answering: A survey.
\newblock {\em arXiv preprint arXiv:2111.10056}, 2021.

\bibitem{Liu2018DelayedIO}
L.~T. Liu, S.~Dean, E.~Rolf, M.~Simchowitz, and M.~Hardt.
\newblock Delayed impact of fair machine learning.
\newblock {\em ArXiv}, abs/1803.04383, 2018.

\bibitem{Liu2020SemiSupervisedMI}
Q.~Liu, L.~Yu, L.~Luo, Q.~Dou, and P.-A. Heng.
\newblock Semi-supervised medical image classification with relation-driven
  self-ensembling model.
\newblock {\em IEEE Transactions on Medical Imaging}, 39:3429--3440, 2020.

\bibitem{lowe2009stride}
H.~J. Lowe, T.~A. Ferris, P.~M. Hernandez, and S.~C. Weber.
\newblock Stride--an integrated standards-based translational research
  informatics platform.
\newblock In {\em AMIA Annual Symposium Proceedings}, volume 2009, page 391.
  American Medical Informatics Association, 2009.

\bibitem{Lu2021FairCP}
C.~Lu, A.~Lemay, K.~Chang, K.~Hoebel, and J.~Kalpathy-Cramer.
\newblock Fair conformal predictors for applications in medical imaging.
\newblock {\em ArXiv}, abs/2109.04392, 2021.

\bibitem{Lu2021EvaluatingSD}
C.~Lu, A.~Lemay, K.~Hoebel, and J.~Kalpathy-Cramer.
\newblock Evaluating subgroup disparity using epistemic uncertainty in
  mammography.
\newblock {\em ArXiv}, abs/2107.02716, 2021.

\bibitem{madras2018learning}
D.~Madras, E.~Creager, T.~Pitassi, and R.~Zemel.
\newblock Learning adversarially fair and transferable representations.
\newblock In {\em International Conference on Machine Learning}, pages
  3384--3393. PMLR, 2018.

\bibitem{mccarty2011emerge}
C.~A. McCarty, R.~L. Chisholm, C.~G. Chute, I.~J. Kullo, G.~P. Jarvik, E.~B.
  Larson, R.~Li, D.~R. Masys, M.~D. Ritchie, D.~M. Roden, et~al.
\newblock The emerge network: a consortium of biorepositories linked to
  electronic medical records data for conducting genomic studies.
\newblock {\em BMC medical genomics}, 4:1--11, 2011.

\bibitem{Mehrabi2021ASO}
N.~Mehrabi, F.~Morstatter, N.~A. Saxena, K.~Lerman, and A.~G. Galstyan.
\newblock A survey on bias and fairness in machine learning.
\newblock {\em ACM Computing Surveys (CSUR)}, 54:1 -- 35, 2021.

\bibitem{Meng2021MIMICIFIA}
C.~Meng, L.~Trinh, N.~Xu, and Y.~Liu.
\newblock Mimic-if: Interpretability and fairness evaluation of deep learning
  models on mimic-iv dataset.
\newblock {\em ArXiv}, abs/2102.06761, 2021.

\bibitem{Minot2021InterpretableBM}
J.~R. Minot, N.~Cheney, M.~E. Maier, D.~C. Elbers, C.~M. Danforth, and P.~S.
  Dodds.
\newblock Interpretable bias mitigation for textual data: Reducing gender bias
  in patient notes while maintaining classification performance.
\newblock {\em ArXiv}, abs/2103.05841, 2021.

\bibitem{Nguyen2019PredictingCR}
M.~Nguyen.
\newblock Predicting cardiovascular risk using electronic health records.
\newblock 2019.

\bibitem{Parikh2019AddressingBI}
R.~B. Parikh, S.~Teeple, and A.~S. Navathe.
\newblock Addressing bias in artificial intelligence in health care.
\newblock {\em JAMA}, 2019.

\bibitem{petersen2010alzheimer}
R.~C. Petersen, P.~S. Aisen, L.~A. Beckett, M.~C. Donohue, A.~C. Gamst, D.~J.
  Harvey, C.~R. Jack, W.~J. Jagust, L.~M. Shaw, A.~W. Toga, et~al.
\newblock Alzheimer's disease neuroimaging initiative (adni): clinical
  characterization.
\newblock {\em Neurology}, 74(3):201--209, 2010.

\bibitem{pfefferbaum2018accelerated}
A.~Pfefferbaum, N.~M. Zahr, S.~A. Sassoon, D.~Kwon, K.~M. Pohl, and E.~V.
  Sullivan.
\newblock Accelerated and premature aging characterizing regional cortical
  volume loss in human immunodeficiency virus infection: contributions from
  alcohol, substance use, and hepatitis c coinfection.
\newblock {\em Biological Psychiatry: Cognitive Neuroscience and Neuroimaging},
  3(10):844--859, 2018.

\bibitem{pfohl2022net}
S.~Pfohl, Y.~Xu, A.~Foryciarz, N.~Ignatiadis, J.~Genkins, and N.~Shah.
\newblock Net benefit, calibration, threshold selection, and training
  objectives for algorithmic fairness in healthcare.
\newblock In {\em Proceedings of the 2022 ACM Conference on Fairness,
  Accountability, and Transparency}, pages 1039--1052, 2022.

\bibitem{Pfohl2019CounterfactualRF}
S.~R. Pfohl, T.~Duan, D.~Y. Ding, and N.~H. Shah.
\newblock Counterfactual reasoning for fair clinical risk prediction.
\newblock {\em ArXiv}, abs/1907.06260, 2019.

\bibitem{Pfohl2019CreatingFM}
S.~R. Pfohl, B.~J. Marafino, A.~Coulet, F.~Rodriguez, L.~P. Palaniappan, and
  N.~H. Shah.
\newblock Creating fair models of atherosclerotic cardiovascular disease risk.
\newblock {\em Proceedings of the 2019 AAAI/ACM Conference on AI, Ethics, and
  Society}, 2019.

\bibitem{pinot2019unified}
R.~Pinot, F.~Yger, C.~Gouy-Pailler, and J.~Atif.
\newblock A unified view on differential privacy and robustness to adversarial
  examples.
\newblock {\em arXiv preprint arXiv:1906.07982}, 2019.

\bibitem{Pleiss2017OnFA}
G.~Pleiss, M.~Raghavan, F.~Wu, J.~M. Kleinberg, and K.~Q. Weinberger.
\newblock On fairness and calibration.
\newblock In {\em NIPS}, 2017.

\bibitem{PuyolAntn2021FairnessIC}
E.~Puyol-Ant{\'o}n, B.~Ruijsink, J.~M. Harana, S.~K. Piechnik, S.~Neubauer,
  S.~E. Petersen, R.~Razavi, P.~J. Chowienczyk, and A.~P. King.
\newblock Fairness in cardiac magnetic resonance imaging: Assessing sex and
  racial bias in deep learning-based segmentation.
\newblock In {\em medRxiv}, 2021.

\bibitem{Radford2019LanguageMA}
A.~Radford, J.~Wu, R.~Child, D.~Luan, D.~Amodei, and I.~Sutskever.
\newblock Language models are unsupervised multitask learners.
\newblock 2019.

\bibitem{Rajkomar2018EnsuringFI}
A.~Rajkomar, M.~Hardt, M.~D. Howell, G.~Corrado, and M.~H. Chin.
\newblock Ensuring fairness in machine learning to advance health equity.
\newblock {\em Annals of Internal Medicine}, 169:866--872, 2018.

\bibitem{rajkomar2018ensuring}
A.~Rajkomar, M.~Hardt, M.~D. Howell, G.~Corrado, and M.~H. Chin.
\newblock Ensuring fairness in machine learning to advance health equity.
\newblock {\em Annals of internal medicine}, 169(12):866--872, 2018.

\bibitem{rajotte2021reducing}
J.-F. Rajotte, S.~Mukherjee, C.~Robinson, A.~Ortiz, C.~West, J.~L. Ferres, and
  R.~T. Ng.
\newblock Reducing bias and increasing utility by federated generative modeling
  of medical images using a centralized adversary.
\newblock {\em arXiv preprint arXiv:2101.07235}, 2021.

\bibitem{ricci2022addressing}
M.~A. Ricci~Lara, R.~Echeveste, and E.~Ferrante.
\newblock Addressing fairness in artificial intelligence for medical imaging.
\newblock {\em nature communications}, 13(1):4581, 2022.

\bibitem{ronneberger2015u}
O.~Ronneberger, P.~Fischer, and T.~Brox.
\newblock U-net: Convolutional networks for biomedical image segmentation.
\newblock In {\em International Conference on Medical image computing and
  computer-assisted intervention}, pages 234--241. Springer, 2015.

\bibitem{Scherrer2012SuperresolutionRT}
B.~Scherrer, A.~Gholipour, and S.~Warfield.
\newblock Super-resolution reconstruction to increase the spatial resolution of
  diffusion weighted images from orthogonal anisotropic acquisitions.
\newblock {\em Medical image analysis}, 16 7:1465--76, 2012.

\bibitem{schrouffdiagnosing}
J.~Schrouff, N.~Harris, O.~Koyejo, I.~Alabdulmohsin, E.~Schnider,
  K.~Opsahl-Ong, A.~Brown, S.~Roy, D.~Mincu, C.~Chen, et~al.
\newblock Diagnosing failures of fairness transfer across distribution shift in
  real-world medical settings: Supplement.

\bibitem{SeyyedKalantari2021CheXclusionFG}
L.~Seyyed-Kalantari, G.~Liu, M.~B.~A. McDermott, and M.~Ghassemi.
\newblock Chexclusion: Fairness gaps in deep chest x-ray classifiers.
\newblock {\em Pacific Symposium on Biocomputing. Pacific Symposium on
  Biocomputing}, 26:232--243, 2021.

\bibitem{seyyed2021underdiagnosis}
L.~Seyyed-Kalantari, H.~Zhang, M.~B. McDermott, I.~Y. Chen, and M.~Ghassemi.
\newblock Underdiagnosis bias of artificial intelligence algorithms applied to
  chest radiographs in under-served patient populations.
\newblock {\em Nature medicine}, 27(12):2176--2182, 2021.

\bibitem{shen2020challenges}
X.~Shen, S.~Ma, P.~Vemuri, and G.~Simon.
\newblock Challenges and opportunities with causal discovery algorithms:
  application to alzheimer’s pathophysiology.
\newblock {\em Scientific reports}, 10(1):2975, 2020.

\bibitem{smith1988using}
J.~W. Smith, J.~E. Everhart, W.~Dickson, W.~C. Knowler, and R.~S. Johannes.
\newblock Using the adap learning algorithm to forecast the onset of diabetes
  mellitus.
\newblock In {\em Proceedings of the annual symposium on computer application
  in medical care}, page 261. American Medical Informatics Association, 1988.

\bibitem{subbaswamy2020development}
A.~Subbaswamy and S.~Saria.
\newblock From development to deployment: dataset shift, causality, and
  shift-stable models in health ai.
\newblock {\em Biostatistics}, 21(2):345--352, 2020.

\bibitem{tahir2023fairness}
A.~Tahir, L.~Cheng, and H.~Liu.
\newblock Fairness through aleatoric uncertainty.
\newblock {\em arXiv preprint arXiv:2304.03646}, 2023.

\bibitem{tibshirani1993introduction}
R.~J. Tibshirani and B.~Efron.
\newblock An introduction to the bootstrap.
\newblock {\em Monographs on statistics and applied probability}, 57(1), 1993.

\bibitem{Tiwald2021RepresentativeF}
P.~Tiwald, A.~Ebert, and D.~Soukup.
\newblock Representative \& fair synthetic data.
\newblock {\em ArXiv}, abs/2104.03007, 2021.

\bibitem{topol2019high}
E.~J. Topol.
\newblock High-performance medicine: the convergence of human and artificial
  intelligence.
\newblock {\em Nature medicine}, 25(1):44--56, 2019.

\bibitem{tsai2021conditional}
Y.-H.~H. Tsai, M.~Q. Ma, H.~Zhao, K.~Zhang, L.-P. Morency, and
  R.~Salakhutdinov.
\newblock Conditional contrastive learning: Removing undesirable information in
  self-supervised representations.
\newblock {\em arXiv preprint arXiv:2106.02866}, 2021.

\bibitem{Tschandl2018TheHD}
P.~Tschandl, C.~Rosendahl, and H.~Kittler.
\newblock The ham10000 dataset, a large collection of multi-source
  dermatoscopic images of common pigmented skin lesions.
\newblock {\em Scientific Data}, 5, 2018.

\bibitem{Wachinger2016DomainAF}
C.~Wachinger and M.~Reuter.
\newblock Domain adaptation for alzheimer's disease diagnostics.
\newblock {\em NeuroImage}, 139:470--479, 2016.

\bibitem{wang2019balanced}
T.~Wang, J.~Zhao, M.~Yatskar, K.-W. Chang, and V.~Ordonez.
\newblock Balanced datasets are not enough: Estimating and mitigating gender
  bias in deep image representations.
\newblock In {\em Proceedings of the IEEE/CVF International Conference on
  Computer Vision}, pages 5310--5319, 2019.

\bibitem{Wang2017ChestXRay8HC}
X.~Wang, Y.~Peng, L.~Lu, Z.~Lu, M.~Bagheri, and R.~M. Summers.
\newblock Chestx-ray8: Hospital-scale chest x-ray database and benchmarks on
  weakly-supervised classification and localization of common thorax diseases.
\newblock {\em 2017 IEEE Conference on Computer Vision and Pattern Recognition
  (CVPR)}, pages 3462--3471, 2017.

\bibitem{wang2022brief}
X.~Wang, Y.~Zhang, and R.~Zhu.
\newblock A brief review on algorithmic fairness.
\newblock {\em Management System Engineering}, 1(1):7, 2022.

\bibitem{Williams2019TowardsQO}
J.~R. Williams and N.~Razavian.
\newblock Towards quantification of bias in machine learning for healthcare: A
  case study of renal failure prediction.
\newblock {\em ArXiv}, abs/1911.07679, 2019.

\bibitem{xu2022algorithmic}
J.~Xu, Y.~Xiao, W.~H. Wang, Y.~Ning, E.~A. Shenkman, J.~Bian, and F.~Wang.
\newblock Algorithmic fairness in computational medicine.
\newblock {\em medRxiv}, 2022.

\bibitem{Xu2022AlgorithmicFI}
J.~N. Xu, Y.~Xiao, W.~Wang, Y.~Ning, E.~A. Shenkman, J.~Bian, and F.~Wang.
\newblock Algorithmic fairness in computational medicine.
\newblock In {\em medRxiv}, 2022.

\bibitem{Xu2014DeepLO}
Y.~Xu, T.~Mo, Q.~Feng, P.~Zhong, M.~Lai, and E.~I.-C. Chang.
\newblock Deep learning of feature representation with multiple instance
  learning for medical image analysis.
\newblock {\em 2014 IEEE International Conference on Acoustics, Speech and
  Signal Processing (ICASSP)}, pages 1626--1630, 2014.

\bibitem{Xue2019RobustLA}
C.~Xue, Q.~Dou, X.~Shi, H.~Chen, and P.-A. Heng.
\newblock Robust learning at noisy labeled medical images: Applied to skin
  lesion classification.
\newblock {\em 2019 IEEE 16th International Symposium on Biomedical Imaging
  (ISBI 2019)}, pages 1280--1283, 2019.

\bibitem{yuan2021assessing}
M.~Yuan, V.~Kumar, M.~A. Ahmad, and A.~Teredesai.
\newblock Assessing fairness in classification parity of machine learning
  models in healthcare.
\newblock {\em arXiv preprint arXiv:2102.03717}, 2021.

\bibitem{Zafar2017FairnessBD}
M.~B. Zafar, I.~Valera, M.~Gomez-Rodriguez, and K.~P. Gummadi.
\newblock Fairness beyond disparate treatment \& disparate impact: Learning
  classification without disparate mistreatment.
\newblock {\em Proceedings of the 26th International Conference on World Wide
  Web}, 2017.

\bibitem{zhang2021recommendation}
D.~Zhang and J.~Wang.
\newblock Recommendation fairness: From static to dynamic.
\newblock {\em arXiv preprint arXiv:2109.03150}, 2021.

\bibitem{Zhang2020HurtfulWQ}
H.~Zhang, A.~X. Lu, M.~Abdalla, M.~B.~A. McDermott, and M.~Ghassemi.
\newblock Hurtful words: quantifying biases in clinical contextual word
  embeddings.
\newblock {\em Proceedings of the ACM Conference on Health, Inference, and
  Learning}, 2020.

\bibitem{zhao2020training}
Q.~Zhao, E.~Adeli, and K.~M. Pohl.
\newblock Training confounder-free deep learning models for medical
  applications.
\newblock {\em Nature communications}, 11(1):6010, 2020.

\end{thebibliography}

\begin{IEEEbiography}[{\includegraphics[width=1in,height=1.25in,clip,keepaspectratio]{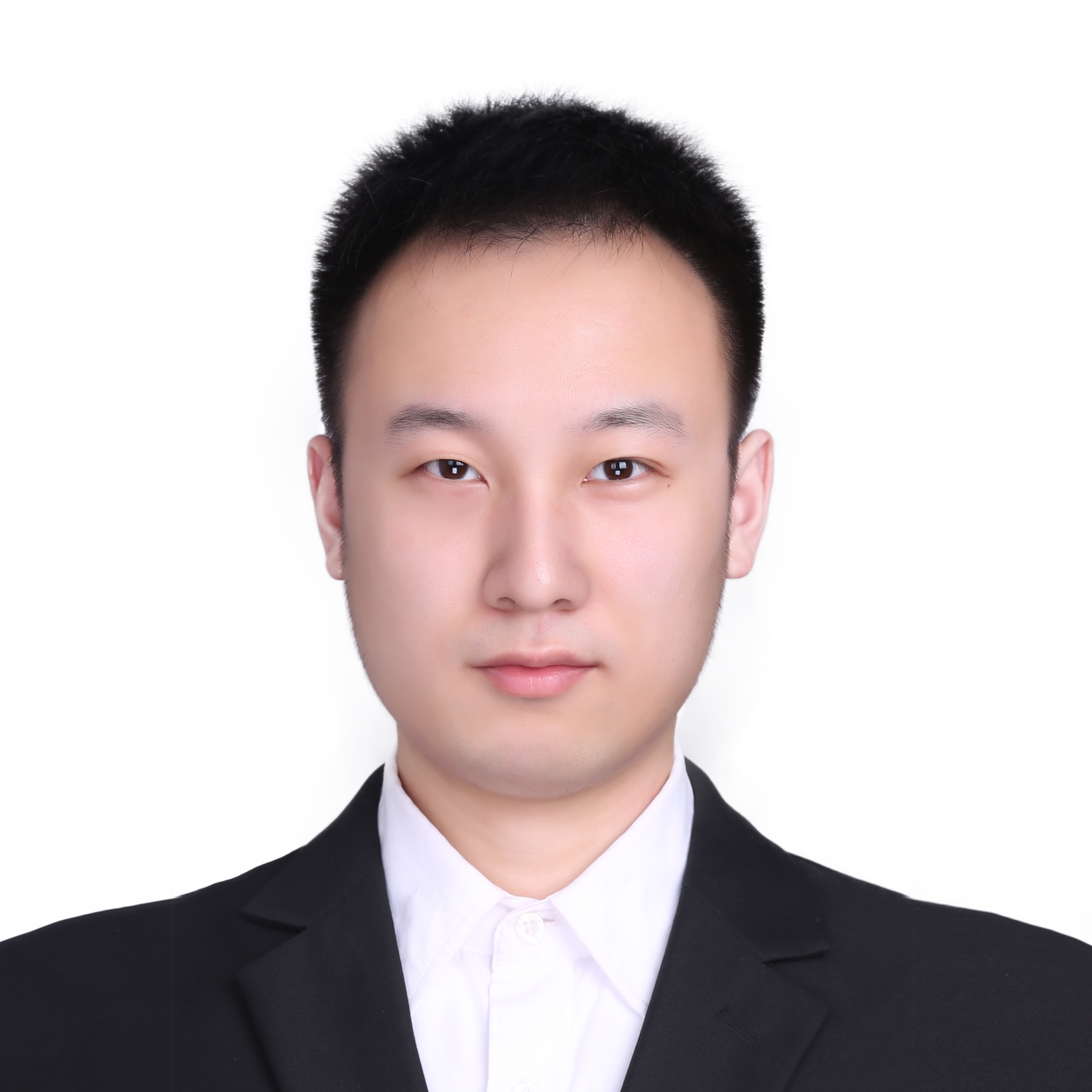}}]%
{Qizhang Feng}{\space} received the B.Eng. degree in Electrical Engineering and Automation from the Huazhong University of Science and Technology, Hubei, China, in 2017, and the master’s degree in Electrical and Computer Engineering from Duke University, NC, USA, in 2020. He is currently working toward the Ph.D. degree in computer engineering with DATA Lab, Texas A\&M University, TX, USA.
His research interests include XAI, machine learning fairness and graph learning.

\end{IEEEbiography}

\begin{IEEEbiography}[{\includegraphics[width=1in,height=1.25in,clip,keepaspectratio]{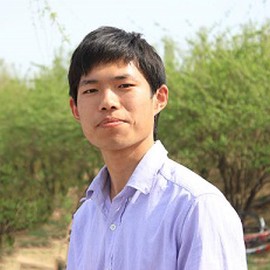}}]%
{Dr. Mengnan Du}{\space} is currently an is an Assistant Professor in the Department of Data Science, New Jersey Institute of Technology (NJIT). Mengnan Du earned his Ph.D. in Computer Science from Texas A\&M University. He has previously worked/interned with Microsoft Research (MSR), Adobe Research, Intel, Baidu Research, Baidu Search Science and JD Explore Academy. His research covers a wide range of trustworthy machine learning topics, such as model explainability, fairness, and robustness. He has had more than 40 papers published in prestigious venues such as NeurIPS, AAAI, KDD, WWW, ICLR, and ICML. He received over 2,300 citations with an H-index of 16. 
\end{IEEEbiography}

\begin{IEEEbiography}[{\includegraphics[width=1in,height=1.25in,clip,keepaspectratio]{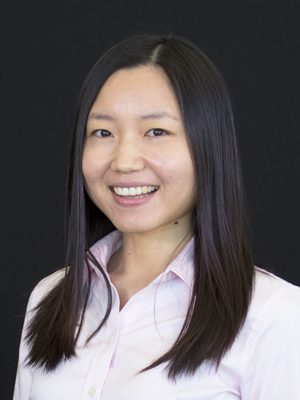}}]%
{Dr. Na Zou}{\space} is currently a Corrie\&Jim Furber ’64 assistant professor in Engineering Technology and Industrial Distribution at Texas A\&M University. She was an Instructional Assistant Professor in Industrial and Systems Engineering at Texas A\&M University from 2016 to 2020. She holds both a Ph.D. in Industrial Engineering and a MSE in Civil, Environmental and Sustainable Engineering from Arizona State University. Her research focuses on fair and interpretable machine learning, transfer learning, network modeling and inference, supported by NSF and industrial sponsors. The research projects have resulted in publications at prestigious journals such as Technometrics, IISE Transactions and ACM Transactions, including one Best Paper Finalist and one Best Student Paper Finalist at INFORMS QSR section and two featured articles at ISE Magazine. She was the recipient of IEEE Irv Kaufman Award and Texas A\&M Institute of Data Science Career Initiation Fellow.
\end{IEEEbiography}

\begin{IEEEbiography}[{\includegraphics[width=1in,height=1.25in,clip,keepaspectratio]{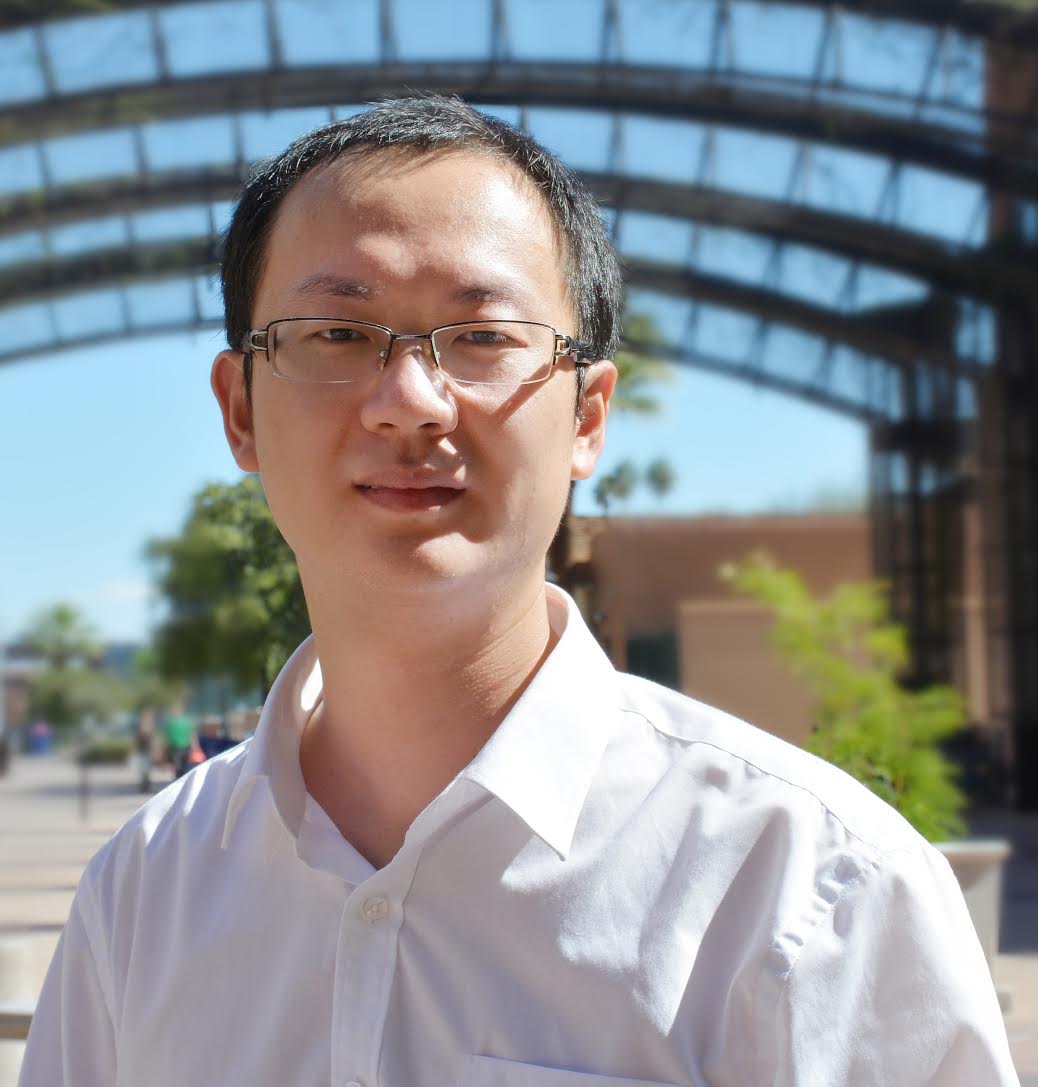}}]%
{Dr. Xia “Ben” Hu} {\space} is an Associate Professor at Rice University in the Department of Computer Science. Dr. Hu has published over 100 papers in several major academic venues, including NeurIPS, ICLR, KDD, WWW, IJCAI, AAAI, etc. An open-source package developed by his group, namely AutoKeras, has become the most used automated deep learning system on Github (with over 8,000 stars and 1,000 forks). Also, his work on deep collaborative filtering, anomaly detection and knowledge graphs have been included in the TensorFlow package, Apple production system and Bing production system, respectively. His papers have received several Best Paper (Candidate) awards from venues such as WWW, WSDM and ICDM. He is the recipient of NSF CAREER Award and ACM SIGKDD Rising Star Award. His work has been cited more than 12,000 times with an h-index of 43. He was the conference General Co-Chair for WSDM 2020. 
\end{IEEEbiography}

\end{document}